\documentclass{article}
\usepackage{PRIMEarxiv}
\errorcontextlines\maxdimen

\usepackage[utf8]{inputenc} % allow utf-8 input
\usepackage[T1]{fontenc}    % use 8-bit T1 fonts
\usepackage{url}            % simple URL typesetting
\usepackage{booktabs}       % professional-quality tables
\usepackage{nicefrac}       % compact symbols for 1/2, etc.
\usepackage{microtype}      % microtypography
\usepackage{fancyhdr}       % header
\usepackage{graphicx}
\usepackage{xcolor}
\usepackage{threeparttable}
\usepackage{booktabs}
\usepackage{multirow}
\usepackage{tabularx}
\usepackage{arydshln}
\usepackage{subcaption}

\usepackage{lineno,hyperref}
\hypersetup{colorlinks, citecolor=blue, linkcolor=red, urlcolor=green}
\usepackage{amsmath,amsfonts,amssymb}
\usepackage{graphics}
\usepackage{bm}
\usepackage{multirow}
\usepackage{algorithm}
\usepackage{algpseudocode}

\usepackage{mathtools}
\usepackage{siunitx}
\DeclarePairedDelimiterXPP\BigOSI[2]%
  {\mathcal{O}}{(}{)}{}%
  {\SI{#1}{#2}}

\title{From Local Interactions to Global Operators: Scalable Gaussian Process Operator for Physical Systems}

\author{Sawan Kumar \\
  Department of Applied Mechanics\\
  Indian Institute of Technology Delhi\\
  \texttt{sawan.kumar@am.iitd.ac.in} \\
  \And
  Tapas Tripura\\
  Department of Applied Mechanics\\
  Indian Institute of Technology Delhi\\
  \texttt{tapas.t@am.iitd.ac.in} \\
  \And
  Rajdip Nayek\\
  Department of Applied Mechanics\\
  Indian Institute of Technology Delhi\\
  \texttt{rajdipn@am.iitd.ac.in} \\
  \And
  Souvik Chakraborty \\
  Department of Applied Mechanics \\
  Yardi School of Artificial Intelligence (ScAI) \\
  Indian Institute of Technology Delhi \\
  The Grainger College of Engineering \\
  University of Illinois Urbana-Champaign \\
  Urbana, IL 61801, USA \\
  \texttt{souvik@am.iitd.ac.in}
}

\begin{document}
\maketitle

\begin{abstract}
Operator learning offers a powerful paradigm for solving parametric partial differential equations (PDEs), but scaling probabilistic neural operators such as the recently proposed Gaussian Processes Operators (GPOs) to high-dimensional, data-intensive regimes remains a significant challenge. In this work, we introduce a novel, scalable GPO, which capitalizes on sparsity, locality, and structural information through judicious kernel design. Addressing the fundamental limitation of cubic computational complexity, our method leverages nearest-neighbor-based local kernel approximations in the spatial domain, sparse kernel approximation in the parameter space, and structured Kronecker factorizations to enable tractable inference on large-scale datasets and high-dimensional input. While local approximations often introduce accuracy trade-offs due to limited kernel interactions, we overcome this by embedding operator-aware kernel structures and employing expressive, task-informed mean functions derived from neural operator architectures. Through extensive evaluations on a broad class of nonlinear PDEs--including Navier-Stokes, wave advection, Darcy flow, and Burgers’ equations--we demonstrate that our framework consistently achieves high accuracy across varying discretization scales. These results underscore the potential of our approach to bridge the gap between scalability and fidelity in GPO, offering a compelling foundation for uncertainty-aware modeling in complex physical systems.

\end{abstract}


\section{Introduction}
Neural operators (NOs) have emerged as a powerful class of deep learning architectures for learning mappings between infinite-dimensional function spaces, offering a paradigm shift in the data-driven solution of partial differential equations (PDEs). Unlike conventional neural networks that operate on fixed-dimensional input-output pairs, NOs are designed to approximate underlying operators, enabling generalization across varying input resolutions and discretizations. This capability has made them especially attractive for scientific and engineering problems governed by complex physics. Recent advances have produced several state-of-the-art architectures -- such as DeepONets \cite{lu2019deeponet, lu2021learning}, Fourier Neural Operators (FNOs) \cite{li2020fourier}, Wavelet Neural Operators (WNOs) \cite{tripura2023wavelet, N2024116546, tripura2024multi, tripura2023foundational}, Laplace Neural Operator (LNO) \cite{cao2023lnolaplaceneuraloperator}, and Variational informed Neural Operator (VINO) \cite{eshaghi2025variational} -- each demonstrating impressive performance across a variety of PDE-based tasks. While these models exhibit high accuracy and scalability, they are inherently deterministic, offering no mechanism to quantify uncertainty. This limitation is particularly critical in high-stakes domains such as medical diagnostics \cite{tripura2023wavelet_elastography}, fracture mechanics \cite{Goswami_2022, kiyani2025predicting}, and weather forecasting \cite{bora2023learningbiascorrectionsclimate, pathak2022fourcastnet, lin2023spherical}, where reliable uncertainty estimates are essential. As such, there is a growing need to augment operator learning frameworks with probabilistic models that can retain the expressive capacity of NOs while providing principled uncertainty quantification.

To overcome the limitations of deterministic neural operators, recent efforts have introduced probabilistic extensions that enable uncertainty quantification within operator learning frameworks. For instance, Bayesian DeepONets \cite{garg2022variational, garg2023randomizedpriorwaveletneural} have demonstrated the feasibility of incorporating uncertainty estimates into operator-based models.  
However, these methods often rely on Monte Carlo sampling or variational inference, which scale poorly with the number of parameters and can become computationally prohibitive in practice. 
In response, Gaussian Process Operators (GPOs) have emerged as a promising alternative, extending the Bayesian framework of Gaussian Processes (GPs) to infinite-dimensional function spaces relevant to operator learning tasks \cite{Magnani2022ApproximateBN, batlle2023kernel, Yang_2021}. GPOs retain the interpretability and principled uncertainty quantification of traditional GPs, while leveraging the structural advantages of neural operators. By incorporating neural architectures into the mean and covariance functions, GPOs can model non-stationary behaviors and learn latent representations that are invariant to input resolution \cite{kumar2025towards}. 
This synergy enhances both expressiveness and generalization. Despite these advantages, GPOs face a fundamental scalability barrier due to the cubic complexity of exact GP inference. To mitigate this in the context of classical GP--based regression, sparse approximation techniques such as inducing points \cite{hensman2013gaussian}, hierarchical models \cite{datta2016hierarchical}, and local kernel approximations based on nearest neighbors \cite{wu2022variational} have been explored. While these methods offer computational relief, they often introduce accuracy trade-offs, particularly in high-dimensional or complex operator learning regimes -- underscoring the need for more effective and scalable GPO formulations.

Several strategies in classical Gaussian processes have been proposed to alleviate this computational bottleneck, notably inducing-point methods and structured covariance approximations such as Kronecker decompositions of kernels \cite{cunningham2023actually,hensman2017VFF,wilson2015kernel, hensman2015scalable,shi2020sparse}. Inducing points effectively reduces inference complexity by summarizing information into a sparse representation, enabling approximate inference with linear complexity in the number of data points. Similarly, Kronecker structured kernels exploit locality and tensor product structures inherent in spatial-temporal datasets, significantly enhancing scalability.
Moreover, recent studies have explored innovative optimization strategies to address the computational bottleneck inherent in the exact inference of GPOs, which cannot be decomposed into mini-batches and are prohibitive for large training datasets.
Stochastic dual descent (SDD) \cite{lin2023stochastic} is a promising approach that leverages a primal-to-dual formulation to recast the inference problem, thereby enabling stochastic optimization that scales more effectively with the number of training samples. While SDD reduces computational complexity with respect to training data, it does not scale with the discretization of input samples -- a critical component for high-resolution operator learning problems. This limitation underscores the need to further enhance the scalability of GPOs. 

Motivated by the limitations discussed above, we introduce a novel \textbf{Lo}cal-\textbf{G}l\textbf{o}bal doubly \textbf{S}parse \textbf{G}aussian \textbf{P}rocess \textbf{O}perator (LoGoS-GPO) that explicitly leverages sparsity, locality, and inherent structure present in data through a hybrid design. Our method integrates inducing point- and local kernel-based approximations with Kronecker-structured covariance factorization, yielding a scalable and expressive probabilistic operator learning model. By combining these structured approximations with operator-informed priors, the framework maintains the expressive power and uncertainty quantification capabilities of full GPs while dramatically reducing computational cost. This design enables efficient training and inference on high-resolution, large-scale datasets, making GPOs viable for practical deployment in scientific computing applications. The key contributions of this work are as follows
\begin{itemize}
    \item \textbf{Hybrid sparsification algorithm:} We develop an efficient sparsification strategy that combines Kronecker-structured kernels, nearest-neighbor local approximations, and inducing point methods to balance training time with predictive accuracy.
    \item \textbf{Principled uncertainty quantification:} Our framework retains the inherent Bayesian nature of GPs, enabling reliable predictive uncertainty estimates essential for decision-making in high-stakes domains.
    \item \textbf{Scalability to large-scale problems:} Through structured and local approximations, the proposed method addresses both sample-level and discretization-level scalability bottlenecks, significantly broadening the applicability of GPOs to real-world PDE-governed systems.
    \item \textbf{Enhanced neural operator prior:} We introduce an improved Wavelet Neural Operator (WNO) architecture within the GPO framework to serve as expressive and informative priors for mean and covariance modeling, further boosting performance on complex operator learning tasks.
\end{itemize}

The subsequent sections of this paper are organized as follows. The review of background concepts related to Gaussian processes is discussed in Section \ref{PS}. Section \ref{sec:framework} explains the proposed framework's mathematical details, including the proposed architecture and details regarding the training and inference. Section \ref{Numercial_eg} demonstrates our approach's implementation on several benchmark examples from the operator learning literature. Finally, Section \ref{Conclusion} provides a concluding review of the results and prospective future work in the study. 

\section{A quick review of Gaussian Process Operator}\label{PS}
In this section, we review the original Gaussian Process Operator (GPO) as proposed in \cite{kumar2025towards}, along with its strengths and weaknesses. This will pave the path for the proposed sparse variant of the GPO discussed in the subsequent sections. However, before proceeding with the discussion of GPO, we quickly formalize the mathematical foundation of the neural operator for the sake of completeness.

\subsection{Operator learning}
Let $\Omega \subset \mathbb{R}^{d_s}$ be a well-defined bounded spatial domain with boundary $\partial\Omega$, where $d_s = 1, 2,$ or $3$ depending on whether the problem is posed in one, two, or three spatial dimensions. We consider a family of differential operators $\mathcal{L} \in \mathcal{C}$ ($\mathcal{C}$ is the continuous space), each acting on a scalar solution function $\hat{u}: \Omega \to \mathbb{R}$, and potentially subject to boundary conditions on $\partial\Omega$. For instance, an elliptic PDE can be written in the general form:
\begin{equation}\label{eq:pde_param}
    \begin{aligned}
    \mathcal{L}[\hat{u}; \lambda](\hat{\bm{x}}) &= 0, &&\quad \hat{\bm{x}} \in \Omega, \\
    \mathcal{B}[\hat{u}](\hat{\bm{x}}) &= 0, &&\quad \hat{\bm{x}} \in \partial \Omega,
    \end{aligned}
\end{equation}
where $\lambda(\hat{\bm{x}})$ denotes input parameters such as spatially varying coefficients or source terms, and $\mathcal{B}$ enforces appropriate boundary conditions.
Consider $\mathcal{A} \in \mathcal{C}$ and $\mathcal{U} \in \mathcal{C}$ to be real separable Banach spaces representing input and output function spaces, respectively. The mapping between these two function spaces can be expressed using the operator $\mathcal{G}^{\dagger}$ as,
\begin{equation}
    \mathcal{G}^{\dagger}: \mathcal{A} \;\longmapsto\; \mathcal{U},
\end{equation}
which maps an input function 
% $\hat{\bm{a}} \in \mathcal{A}$ 
\(\mathcal{A} \owns \hat{\bm{a}}: \Omega_{a} \mapsto \mathbb{R}^{d_a^{'}}\) to the corresponding unique solution $\mathcal{U} \owns \hat{\bm{u}}: \Omega_u \mapsto \mathbb{R}^{d_u^{'}}$. 
We observe $N$ input--output function pairs $\{(\hat{\bm{a}}_1(\bm x_a), \hat{\bm{u}}_1(\bm x_u)), \ldots, (\hat{\bm{a}}_N(\bm x_a), \hat{\bm{u}}_N(\bm x_u))\}$,
where every 
$\hat{\bm{u}}_i(\bm x_u) = \mathcal{G}^{\dagger}[\hat{\bm{a}}_i](\bm x_a)$. 
In practice, the continuous functions $\{\hat{\bm{a}}_i,\hat{\bm{u}}_i\}_{i=1}^{N} \;\subset\; \mathcal{A} \times \mathcal{U}$ are only available at discrete spatial locations $\bm x_a \in \Omega_{a,\textit{grid}} \subset \mathbb{R}^{d_a}$ and $\bm x_u \in \Omega_{u,\textit{grid}} \subset \mathbb{R}^{d_u}$, yielding finite-dimensional approximations, $\bm a_i = \hat{\bm{a}}_i(\bm x_a) \in \mathbb{R}^{d_a}$ and $\bm{u}_i = \hat{\bm{u}}_i(\bm x_u) \in \mathbb{R}^{d_u}$, where $d_a, d_u$ denotes the number of discrete points in their respective domains. 
For simplicity, and without the loss of generality, we consider the domains \(\Omega_a = \Omega_u\) and the same discretization for both the input and output functions that is \( \bm{x}_a = \bm{x}_u = \bm x \in \Omega_{\textit{grid}} \; \subset \mathbb{R}^d\).
The objective in operator learning is to approximate the true, potentially intractable operator \(\mathcal{G}^\dagger\) using a finite set of observed input-output function pairs. To this end, we seek a parameterized operator \(\mathcal{G}\) such that,
\begin{equation}
    \mathcal{G} : \mathcal{A} \times \mathbf{\Theta} \;\longmapsto\; \mathcal{U},
\end{equation}
where \(\mathbf{\Theta} \subset \mathbb{R}^m\) denotes a finite-dimensional set of learnable parameters. The goal is to learn \(\mathcal{G}\) such that it generalizes well across the function space \(\mathcal{A}\), thereby approximating \(\mathcal{G}^\dagger\) in a data-driven manner.

\subsection{Gaussian Process Operator}\label{subsec:gpo}
In Gaussian Process Operator (GPO) \cite{kumar2025towards}, the goal is to model the unknown solution operator \(\mathcal{G}\) using a Gaussian process prior. For a given input function \(\hat{\bm a} \in \mathcal{A}\), the output \(\hat{\bm u} \in \mathcal{U}\) can be represented as
\begin{equation}
    \hat{\bm{u}}(\bm x)\;=\; \mathcal{G}[\hat{\bm{a}}; \bm \theta](\bm x),
\end{equation}
where $\bm \theta$ are the learnable parameters that parametrizes the operator \(\mathcal{G}\). To learn this mapping, GPO considers a function-valued GP prior over $\mathcal{G}$, given as,
\begin{equation}
  \mathcal{G} \;\sim\; \mathcal{GP}\Bigl(\mathcal{M},\; \mathcal{K}_o \Bigr),
  \label{eq:gp_prior_1}
\end{equation}
where \(\mathcal{M}:\mathcal{A} \mapsto \mathcal{U}\) is the mean operator, which is commonly set to zero for simplicity. The term \(\mathcal{K}_o\) represents an operator-valued covariance kernel that maps pairs of input functions to bounded linear operators in \(\mathrm{Lin}(\mathcal{U})\), the space of linear maps acting on the output function space \(\mathcal{U}\) \cite{kadri2016operator,minh2016operator}.

In practice, dealing with infinite-dimensional function space requires discretizing both the input and output functions over a finite grid. This leads to the challenge of linking the underlying function-valued Gaussian process with its finite-dimensional representation. We exploit the \emph{probabilistic currying theorem} \cite{magnani2024linearizationturnsneuraloperators} to establish the equivalence between the function-valued and vector-valued GPs  \cite{zio2022multi,watanabe2025derivationoutputcorrelationinferences} over an augmented input space,
\begin{equation}
\mathcal{G}(\bm{a})(\bm x) \overset{\text{a.s.}}{=} \bm{f}(\bm a , \bm x),
\end{equation}
where \(\bm{f} : \mathbb{R}^{d_a} \times \Omega_{\textit{grid}} \mapsto \mathbb{R}^{d_u}\) is a finite-dimensional vector-valued function that can be modeled using a multi-output GP prior with zero mean as considered in the original GPO implementation \cite{kumar2025towards}, 
\begin{equation}
    \bm{f}(\bm a, \bm x) \sim \mathcal{GP}\Bigl(\bm 0,\; \bm{k} \bigl((\bm a,\bm x),(\bm a', \bm x') ;\,\bm {\theta}_k\bigr)\Bigr)
\end{equation}
where \(\bm{f}(\bm a,\bm x) \in \mathbb{R}^{d_u}\) is the vector-valued output, the mean taken as zero for simplicity. The term \(\bm{k} : \mathbb{R}^{d_a + d} \times \mathbb{R}^{d_a + d} \mapsto \mathbb{R}^{(d_u \times d_u)}\) is a multi-output covariance kernel \cite{alvarez2012kernels} parameterized by \(\bm{\theta}_k\),  which typically includes hyperparameters such as lengthscales and signal variances. Note that this established equivalence in \cite{magnani2024linearizationturnsneuraloperators} facilitates the conceptualization of computationally tractable real-valued Gaussian processes as infinite-dimensional, function-valued entities.

Further in GPO, the covariance kernel is constructed in a transformed latent space by projecting the inputs through a neural operator. Instead of applying a conventional kernel (e.g., Matérn \cite{gneiting2006geostatistical} or RBF \cite{Rasmussen2004}) directly in the original input space, a mapping of the following form is used,
\begin{equation}\label{eq:latent_emb}
    \Phi_{\bm \theta_{k_1}}: \mathcal{A} \mapsto \mathcal{H}.
\end{equation}
Here $\mathcal{A}$ is the original input space and $\mathcal{H}$ is a learned latent space, where $\mathcal{H}$ is complete in Hilbert space. The covariance kernel is then defined as,
\begin{equation}\label{eq:kappa_F}
    \bm{k}((\bm a,\bm x),(\bm a', \bm x'); \bm \theta_k) = \bm{k}_{ax} \big(\Phi_{\bm \theta_{k_1}}(\bm a,\bm x), \Phi_{\bm \theta_{k_1}}(\bm a', \bm x'); \bm \theta_{k_2}\big),
\end{equation}
where $\bm{k}_{ax}$ is a standard positive-definite kernel applied in the latent space with \(\bm \theta_k = \{\bm \theta_{k_1}, \bm \theta_{k_2}\}\) being the trainable hyperparameters of GPO. This formulation allows the model to capture complex dependencies in the data as it is deeper than the shallow vanilla kernels. However, training GPO is computationally expensive. 
In the original GPO implementation \cite{kumar2025towards}, stochastic dual descent was exploited to address the computational challenge to a certain extent; however, it fails to handle high-dimensional inputs.
Accordingly, the objective of this paper is to develop a sparse variant of the GPO that can scale seamlessly with the number of training samples and input discretization.
\section{Proposed methodology}\label{sec:framework}
In this section, we elaborate on our proposed LoGoS-GPO framework that exploits localized kernels, stochastic variational approximations, and kernel decomposition using Kronecker products for a family of structured and separable kernels. Specifically, we outline the developed LoGoS-GPO in detail, along with associated algorithms. We also provide details on the enhanced WNO, an improved WNO architecture used with the proposed LoGoS-GPO.

\subsection{LoGoS-GPO}\label{Framework}
The proposed \textbf{Lo}cal-\textbf{G}l\textbf{o}bal doubly \textbf{S}parse \textbf{G}aussian \textbf{P}rocess \textbf{O}perator (LoGoS-GPO) builds on the GPO discussed in Section \ref{subsec:gpo}. 
However, unlike the original implementation, LoGoS-GPO provides a more generalized formulation over GPO by considering a non-zero mean in the Gaussian process. This is the first novelty of the proposed framework. The underlying GP is expressed as follows,
\begin{equation}
  \bm f(\bm a,\bm x) \;\sim\; \mathcal{GP}\Bigl(\bm  m(\bm a,\bm x);\,\bm \theta_m),\; \bm{k}((\bm a,\bm x), (\bm a',\bm x'); \bm \theta_k)\Bigr),
  \label{eq:gp_prior}
\end{equation}
where $\bm{k}((\bm a,\bm x), (\bm a',\bm x'); \bm \theta_k)$ is previously defined in Eq. \eqref{eq:kappa_F}.
As previously noted in \cite{kumar2024neuraloperatorinducedgaussian}, the inclusion of a non-stationary mean also allows the underlying GP to model a non-stationary process using a stationary kernel.
Overall, this is expected to enhance the expressive capability of the existing GPO, which in turn will account for the approximation introduced due to sparsification. We leverage a neural operator to parameterize the mean function $\bm m((\bm a,\bm x);\,\bm \theta_m)$ as follows:
\begin{equation}\label{eq:NO_mean}
    \bm m((\bm a,\bm x);\,\bm \theta_m) = \mathcal{M}[\bm a;\bm \theta_m](\bm x),
\end{equation}
where $\mathcal{M}[\cdot ;\bm \theta_m]$ is a neural operator with trainable parameters \(\bm \theta_m\).
The main computational challenge of GPO arises from the need to invert the dense covariance matrix of cubic time complexity during the naive implementation of exact inference. To mitigate this, we introduce decomposition of the kernel matrix by exploiting locality, inducing points, and Kronecker decomposition, which reduces the cubic computational cost. 
To that end, we first employ \emph{separable Kronecker decomposition}
to decompose the kernel in the spatial and feature space  for the given training set \(\{(\bm{{a}}_i, \bm x), \bm u_i\}_i^{N}\) as follows,
\begin{equation}\label{eq:kf_kernel}
\mathbf{K}
\;=\;
\mathbf{K}_{a}(\Phi_{\bm\theta_{k_3}}(\mathbf{{A}}), \Phi_{\bm\theta_{k_3}}(\mathbf{{A}}'); \bm{\theta}_{a}) \otimes \mathbf{K}_x(\bm{x}, \bm{x}'; \bm{\theta}_x)
\end{equation}
where 
\(\mathbf{A} = \{\bm a_i\}_{i=1}^N\) and
\(\mathbf{K} \in \mathbb{R}^{Nd \times Nd}\) is the covariance matrix formulated using the covariance kernel \(\bm k(\cdot,\cdot)\) in Eq.~\eqref{eq:gp_prior} and the training set \(\{(\bm{{a}}_i, \bm x), \bm u_i\}_{i=1}^{N}\). \(\bm{\theta}_k = \{\bm {\theta}_{k_3},\bm{\theta}_a, \bm{\theta}_x\}\) denotes the parameters of the GPO's covariance kernel and $\otimes$ denotes the Kronecker product.
% The GPO's kernel parameter \(\bm{\theta}_k = \{\bm {\theta}_{k_3},\bm{\theta}_a, \bm{\theta}_x\}\), $\otimes$ denotes
The kernel matrix $\mathbf{K}_{a}(\Phi_{\bm\theta_{k_3}}(\mathbf{{A}}), \Phi_{\bm\theta_{k_3}}(\mathbf{{A}}') \in \,\mathbb{R}^{N\times N}$ is computed using the latent embedding layer \(\Phi_{\bm \theta_{k_3}}(\cdot)\), parameterized by \(\bm \theta_{k_3}\). The covariance matrix $\mathbf{K}_x(\bm{x}, \bm{x}'; \bm{\theta}_x)\in \mathbb{R}^{d\times d}$, captures the covariance on the spatial grid \(\bm x\). The decomposition defined in Eq.~\eqref{eq:kf_kernel} provides a separable covariance structure by utilizing the Kronecker product of covariance matrices. This separable construction significantly reduces computational cost as the covariance matrix factorizes into smaller blocks. When the spatial domain \(\Omega_{\textit{grid}}\) is structured, the computational cost can be further reduced by decomposing the spatial kernel using the Kronecker product.

To further reduce the computational cost, we employ  K-Nearest Neighbors (KNN) to build a sparse kernel capturing the spatial correlation.
Formally, we denote by $\mathcal{S}_{\textit{KNN}}$ the set of the $K \in \mathbb{N}$ nearest neighbors of grid‐point $\bm x$. With the introduction of the neighboring points, we define the sparse spatial kernel as,
\begin{equation}\label{eq:NN_kernel}
    [\mathbf{K}^{N}_x]_{ij}
    \;=\;
    \begin{cases}
    {k} (\bm x_i,\,\bm x_j;\,\bm\theta_{s}), 
    & \text{if } \bm x_j \in \mathcal{S}_{\textit{KNN}},\\[6pt]
    0, 
    & \text{otherwise},
    \end{cases}
\end{equation}
where ${k}(\cdot,\cdot)$ can be defined using any standard kernel (e.g., RBF or Matérn). 
This KNN approximation yields a sparse $d\times d$ covariance matrix with at most \(K \ll d \) non-zeros per row, reducing the computational cost and memory requirements.
Considering \(\{\bm x_i\}_{i=1}^d\) to be the grid points, we compute \(K\) nearest neighbors for each  \(\bm x_i\), 
\begin{equation}
\mathcal{S}
=\bigl\{\bm x_j : j\in\mathrm{arg. \,top \; K}\bigl(-\|\bm x_i-\bm x_j\|\bigr)\bigr\}.
\end{equation}
Here we use the notation \(\mathrm{arg. \, top \; K}\bigl(-\|\bm x_i-\bm x_j\|\bigr)\) to refer the indices of \(K\) smallest Euclidean distance between points \(\bm x_i\) and all the other points \(\bm x_j\).
The implementation steps of the nearest-neighbor sparse spatial covariance kernel is illustrated in the Algorithm \ref{algo:knn}. For \(K\) nearest neighbor points, \(N\) training samples, and \(d\) discretization points,
\begin{algorithm}[!ht]
\caption{Nearest‐neighbor Sparse Spatial Covariance Matrix Construction}
\label{algo:knn}
\begin{algorithmic}[1]
\Require Grid points \(\{\bm x_i\}_{i=1}^d\), neighbor count \(K\), base kernel \({k}(\cdot,\cdot)\)
\Ensure Sparse kernel matrix \( \mathbf{K}^{N}_x\)
\For{\(i=1\) to \(d\)}
  \State Compute distances \(d_{ij}^s = \|\bm x_i - \bm x_j\|\) for \(j=1,\dots,d\)
  \State Find index set \(\mathcal{S}\) of the \(K\) smallest \(d_{ij}^s\)
  \For{\(\bm x_j\in \mathcal{S}\)}
    \State \([\mathbf{K}^{N}_x]_{ij} \gets {k}(\bm x_i,\bm x_j;\,\bm\theta_s)\)
  \EndFor
\EndFor
\end{algorithmic}
\end{algorithm}
the proposed approach requires \(\mathcal{O}(d \log N + d K)\) inversion cost and \(\mathcal{O}(d K)\) memory for \(\mathbf{K}_{\bm x}^N\), as opposed to \(\mathcal{O}(d^3)\) inversion cost and \(\mathcal{O}(d^2)\) memory for the full kernel \(\mathbf{K}_{\bm x}\).

As the third layer of refinement, we employ the sparse variational approximation (SVA) \cite{hensman2013gaussian} on the feature space kernel $\mathbf{K}_{a}(\Phi_{\bm\theta_{k_3}}(\bm{a}), \Phi_{\bm\theta_{k_3}}(\bm{a}'); \bm{\theta}_{a})$ to further enhance the scalability. 
As defined earlier, we have the dataset \( \mathcal{D} = \{\bar {\bm a}_i,\bm u_i\}_{i=1}^{N} \), where the pair $\bm{\bar{a}} = (\bm a_i, \bm x) \in \mathbb{R}^{d\times 2}$ constructs the input to the LoGoS-GP. 
We define the input data matrix as \(\mathbf{\bar{A}} = \{(\bm a_1, \bm x), \dots, (\bm a_N, \bm x)\}^T \in \mathbb{R}^{N \times d \times 2}\), and the output data matrix as \(\mathbf U = \{\bm u_1, \dots, \bm u_N\}^T \in \mathbb{R}^{N \times d}\).
Under the SVA \cite{hensman2013gaussian}, we select a set of \(M\) \emph{inducing points} \(\mathbf{Z}= \{\bm{z}_1,\dots,\bm{z}_M\}\), where \(\bm z_m \in \mathbb R^d \) with \(M \ll N\). 
We define the corresponding latent inducing variables \(\bm{v} = \{\bm{v}_1, \dots , \bm{v}_M\}^T \),  where \(\bm v_m \in \mathbb{R}^d\). Let  \( q(\bm{v}) \) denote the joint variational distribution of \(\bm v\) . Further,  we define the latent function evaluations at the training points as \(\mathbf  F = \{\bm f_i\}_{i=1}^N \), where \(\bm f_i \in \mathbb{R}^d\) are the noise-free evaluations. We also define \(\bm f = \text{vec}(\mathbf F) \in \mathbb{R}^{Nd}\) as the vectorized latent evaluations. Instead of directly approximating the full posterior \( p \bigl(\bm f \mid \mathcal{D}\bigr) \), we define the approximate posterior as
\begin{equation}
  q\bigl(\bm f\bigr)
   =
  \int
  p \bigl(\bm f \;\big|\; \bm{v}\bigr)\,
  q \bigl(\bm{v}\bigr)\,
  \mathrm{d}\bm{v},
\end{equation}
where \( p \bigl(\bm f \;\big|\; \bm{v}\bigr) \) is Gaussian with a mean and covariance determined by the kernel matrix blocks. 
We represent the blocks of the covariance matrix as follows:
\begin{equation}
\begin{aligned}
\mathbf{K}_{ZZ} &= \mathbf{K}_{a}(\Phi_{\bm\theta_{k_3}}(\mathbf{Z}), \Phi_{\bm\theta_{k_3}}(\mathbf{Z}')) \; \otimes \; \mathbf{K}_{x}(\bm{x}, \bm{x}') \in \mathbb{R}^{Md \times Md}, \\
\mathbf{K}_{\bar{A}Z} &= \mathbf{K}_{a}(\Phi_{\bm\theta_{k_3}}(\mathbf{A}), \Phi_{\bm\theta_{k_3}}(\mathbf{Z})) \; \otimes \; \mathbf{K}_{x}(\bm{x}, \bm{x}') \in \mathbb{R}^{Nd \times Md}, \\
\mathbf{K}_{\bar{A}\bar{A}} &= \mathbf{K}_{a}(\Phi_{\bm\theta_{k_3}}(\mathbf{A}), \Phi_{\bm\theta_{k_3}}(\mathbf{A}')) \; \otimes \; \mathbf{K}_{x}(\bm{x}, \bm{x}') \in \mathbb{R}^{Nd \times Nd}.
\end{aligned}
\end{equation}
% where the matrix \(\mathbf{A} = \{\bm a_i\}_{i=1}^N\). 
Note that we never explicitly deal with these huge matrices; instead, we exploit the Kronecker product properties of matrices for efficient computations.

The vectorized forms of the mean are defined as
\begin{equation} \label{eq:mean_Z}
  \bm m_{0}^{(Z)} = \text{vec}(\bm m_0(\mathbf{Z})) \in \mathbb{R}^{Md},  
\end{equation}
and 
\begin{equation}\label{eq:mean_A}
    \bm m_{0}^{(A)} = \text{vec}(\bm m_0(\mathbf{\bar{A}})) \in \mathbb{R}^{Nd}.
\end{equation}
The conditional distribution \(p\bigl(\bm{f} \mid \bm{v}\bigr)\)is given by:
\begin{equation}
\begin{aligned}
p\bigl(\bm{f} \mid \bm{v}\bigr) &= \mathcal{N}(\bm{\mu}, \bm{\Sigma}),\\ \text{with,} \quad 
\bm{\mu} &= \bm{m}_{0}^{(\bar A)} 
+ \mathbf{K}_{\bar{A}Z} 
\left(\mathbf{K}_{ZZ}\right)^{-1} 
(\bm{v} - \bm{m}_{0}^{(Z)}) 
, \\
\bm{\Sigma} &= \mathbf{K}_{\bar{A}\bar{A}} 
- \mathbf{K}_{\bar{A}Z} 
\left(\mathbf{K}_{ZZ}\right)^{-1} 
\mathbf{K}_{Z\bar{A}},
\end{aligned}
\end{equation}
where
\begin{equation}
    \left(\mathbf{K}_{ZZ}\right)^{-1}  = \left[\mathbf{K}_{a}(\Phi_{\bm\theta_{k_3}}(\mathbf{Z}), \Phi_{\bm\theta_{k_3}}(\mathbf{Z}'))\right]^{-1} \; \otimes \; \left[\mathbf{K}_{x}(\bm{x}, \bm{x}')\right]^{-1}.
\end{equation}
We note that the spatial covariance matrix $\mathbf{K}_{x}(\bm{x}, \bm{x}')$ is already sparsified using the K-nearest neighbor scheme as discussed in Algorithm \ref{algo:knn}.
The variational distribution \(q(\bm{v}) \) is typically chosen to be a multivariate normal \(\mathcal{N}\bigl(\bm{m},\,\mathbf{S}\bigr)\) with learnable variational parameters \(\bm{m}\) and \(\mathbf{S}\).
To optimize these parameters, we maximize the evidence lower bound (ELBO), defined as
\begin{equation}\label{eq:Loss_elbo}
\mathcal{L}_{\mathrm{SVGP}} = \sum_{i=1}^N \mathbb{E}_{q(\bm f_i)} \left[ \log p(\bm u_i \mid \bm f_i) \right] - \mathrm{KL}(q(\bm v) \mid\mid p(\bm v)),
\end{equation}
The prior distribution \( p(\bm{v}) \) is induced by the GP prior. Once the optimized parameters are obtained, including \(\bm{m}\) and \(\mathbf{S}\), we can predict responses at a new test input, \(\bar{\bm a}_* = (\bm a_*, \bm x_*) \in \mathbb{R}^{d\times 2}\), by computing: 
\begin{equation}
q\bigl(\bm f_*\bigr)
\;=\;
\int
p\bigl(\bm f_*\,\bigm|\,\bm{v}\bigr)\,
q(\bm{v})
\,d\bm{v}.
\end{equation}
This resulting predictive distribution \(q \left(\bm f_*\right)\) remains Gaussian, and its mean and variance can be derived in closed form,
\begin{subequations}\label{eq:pred_moments}
\begin{align}
  \mathbb{E}_{q}\!\bigl[\bm f(\bar{\bm a}_*)\bigr]
  &=
  \bm m_0(\bar{\bm a}_*)
  \;+\;
  \mathbf K_{*Z}\mathbf{ \left({K_{ZZ}}\right)}^{-1}
  \bigl(\,\bm m - \bm m_0(\bm Z)\bigr),
  \label{eq:pred_mean}
  \\[4pt]
  \mathrm{Var}_{q}\!\bigl[\bm f(\bar{\bm a}_*)\bigr]
  &=
  \mathbf{K}_{**}
  \;-\;
  \mathbf K_{*Z}\mathbf{ \left({K_{ZZ}}\right)}^{-1}
  \bigl(\mathbf K_{ZZ}-\bm S\bigr)
  \mathbf{ \left({K_{ZZ}}\right)}^{-1}\mathbf K_{Z*},
\end{align}
\end{subequations}
where we have kernel matrices \(\mathbf{K}_{*Z} = \mathbf{K}_{a}(\Phi_{\bm\theta_{k_3}}({\bm a}_*), \Phi_{\bm\theta_{k_3}}(\mathbf{Z})) \otimes \mathbf{K}_{x}({\bm x}_*, \bm{x}_*) \in \mathbb{R}^{d \times Md}\), \(\mathbf{K}_{Z*} = \left(\mathbf{K}_{*Z}\right)^T \), \(\mathbf{K}_{**} = \mathbf{K}_{a}(\Phi_{\bm\theta_{k_3}}({\bm a}_*), \Phi_{\bm\theta_{k_3}}({\bm a}_*)) \otimes \mathbf{K}_{x}({\bm x}_*,{\bm x}_*) \in \mathbb{R}^{d \times d}\), and \(\bm m_0(\bm{\bar{a}_*}) \in \mathbb{R}^d\).
Overall, this combination of Kronecker product-based representation, the K-nearest neighbor-based localized representation of the spatial coordinate, and the sparse variational approximation of the covariance kernel
results in an efficient and scalable GPO which reduces the computational cost to \(\mathcal{O}(M^3 + M^2B + dK^2)\) compared to the exact inference implementation of \(\mathcal{O}(N^3d^3)\), where \(N\) is the number of training samples, \(d\) denotes the discretization grid, \(M \) is the number of inducing points, \(K\) represents the nearest-neighbors, and \(B \) is the mini-batch size. The use of neural operator-based mean function with latent-space embedded kernel design, on the other hand, ensures that the proposed algorithm is highly accurate.
The complete training procedure, including the computation of the mean function, kernel matrix, variational objective, and optimization routine, is detailed in Algorithm~\ref{alg:training}. Once the model is trained, one can compute the predictive distribution for a new test input using Algorithm~\ref{alg:inference}. 
\begin{algorithm}[ht!]
    \caption{Training of the framework}
    \label{alg:training}
    \begin{algorithmic}[1]
    
    \Require 
    Training data $\mathcal{D} = \bigl\{(\bar{\bm{a}}_i, \bm{u}_i)\bigr\}_{i=1}^N$ with $\bar{\bm{a}}_i = (\bm{a}_i,\bm{x}_i)$;  
    Neural operator as mean function \(\bm m_0(\cdot;\bm \theta_m)\) with parameters $\bm{\theta}_m$;  number of inducing points $M$ and their locations $\bm{Z} = \{\bm{z}_m\}_{m=1}^{M}$;
    kernel latent embedding \(\Phi(\cdot)\) ,parameters $\bm{\theta}_\Phi$; 
    variational parameters $(\bm{m}, \mathbf{S})$ for $q(\bm{v})$;
    total epochs $T$;
    mini-batch size $B$;
    learning rate $\eta$;
    optimizer (AdamW or L-BFGS)
    \Ensure 
    $\Theta^{(T)} = \{\bm{\theta}_m,\;\bm{\theta}_\Phi,\;\bm{Z},\;\bm{m},\;\mathbf{S}\}$
    
    \State \textbf{Initialize} $\Theta^{(0)}$ \\
    \quad $\Theta \;=\; \bigl\{\bm{\theta}_m,\;\bm{\theta}_\Phi,\;\bm{Z},\;\bm{m},\;\mathbf{S}\bigr\}.$
    
    \For{$e = 1$ to $T$} \Comment{Epoch index}
    
      \State Mini-batch selection:
        Sample $\mathcal{B} \subset \{1,\dots,N\}$ of size $B$.
    
      \State Neural Operator as mean:\\
        \qquad $\bm{m}_{0}^{\mathcal{B}} \;\gets\; \bm m_0 \bigl(\bar{\bm{a}}_{\mathcal{B}};\,\bm{\theta}_m^{(e-1)}\bigr)$, \Comment{Eq.~\eqref{eq:NO_mean}}
            
      \State Latent embedding:
        For each $i\in\mathcal{B}$, compute 
        \(\Phi\!\bigl(\bm{a}_i;\,\bm{\theta}_\Phi^{(e-1)}\bigr)\). \Comment{Eq.~\eqref{eq:latent_emb}}
    
      \State Nearest-neighbor approximation:\\ 
        \qquad $
        {k}_x(\bm{x}_i,\bm{x}_j) 
        \;\approx\; 
        \begin{cases}
        {k}_x(\bm{x}_i,\bm{x}_j), & \text{if } \bm{x}_j \in \mathcal{S},\\
        0, & \text{otherwise}.
        \end{cases}
        $ \Comment{Eq.~\eqref{eq:NN_kernel}}
      \State Form kernel for embedded features: \\ 
        \qquad $ \mathbf{K}_{a}(\Phi_{\bm\theta_{k_3}}(\mathbf{{A}}), \Phi_{\bm\theta_{k_3}}(\mathbf{{A}}'); \bm{\theta}_{a})
                $ \Comment{Eq.~\eqref{eq:kf_kernel}}

      \State ELBO:\\
        \qquad $\mathcal{L}_{\mathrm{SVGP}}(\mathcal{B})
        \;=\;
        \sum_{i \in B}
        \mathbb{E}_{q(f(\bar{\bm{a}}_i))}\!\bigl[
          \log p\bigl(\bm{u}_i \mid f(\bar{\bm{a}}_i)\bigr)
        \bigr]
        \;-\;
        \mathrm{KL}\bigl(q(\bm{v})\;\|\;p(\bm{v})\bigr),$ \\
        \qquad where $q(\bm{v}) = \mathcal{N}\!\bigl(\bm{m},\mathbf{S}\bigr)$. \Comment{Eq.~\eqref{eq:Loss_elbo}}
    
      \State Parameter update:\\
        \qquad $
          \Theta^{(e)} 
          \;\gets\;
          \Theta^{(e-1)} 
          \;-\;
          \eta\nabla_{\Theta}\mathcal{L}_{\mathrm{SVGP}}(\mathcal{B}).
        $
    
    \EndFor
    \State \textbf{Return} $\Theta^{(T)} = \{\bm{\theta}_m,\;\bm{\theta}_\Phi,\;\bm{Z},\;\bm{m},\;\mathbf{S}\}$.
    
    \end{algorithmic}
\end{algorithm}

\subsection{Enhanced wavelet neural operator}\label{subsec:ewno}
One of the features of the proposed LoGoS-GPO is the use of a neural operator to capture the mean function and construct the covariance kernel in the latent space. While any discretization-invariant neural operator can be used within the proposed LoGoS-GPO framework, we utilize the wavelet neural operator in this work \cite{tripura2023wavelet}.
The core idea behind the WNO architecture is first to lift the input function into a high-dimensional space, then apply a sequence of wavelet kernel integral blocks. The global convolution mechanism in the kernel integral blocks learn the underlying operator mapping. Finally, the output is reconstructed by projecting the transformed features back to the original space \cite{tripura2023wavelet}.

Let $\xi: \Omega \to \mathbb{R}$ be an input function defined on a spatial domain $\Omega \subset \mathbb{R}^{d_s}$. Let the forward and inverse wavelet transforms be denoted by $\mathcal{W}$ and $\mathcal{W}^{-1}$, respectively. 
Given the orthonormal mother wavelet $g \in L^2(\mathbb{R})$,these transforms are defined as:
\begin{subequations}
\begin{align}
(\mathcal{W}\xi)(s, t) &= \int_{\Omega} \xi(x) \frac{1}{|s|^{1/2}} g\left(\frac{x - t}{s}\right) dx, \\
(\mathcal{W}^{-1}\xi_w)(x) &= \frac{1}{C_g} \int_0^\infty \int_{\Omega} \xi_w(s, t) \frac{1}{|s|^{1/2}} \tilde{g}\left(\frac{x - t}{s}\right) dt \frac{ds}{s^2},
\end{align}
\end{subequations}
where $\tilde{g}$ is the dual form of the mother wavelet $g$, and $C_g$ is a normalization constant ensuring invertibility, defined as,
\begin{equation}
    \begin{aligned}
        C_g &= 2\pi \int_{\Omega} \frac{1}{|\omega|} |g(\omega)| |\tilde{g}(\omega)|  d\omega, \\
\text{with \;\;} & g^*(\omega) = \int_{-\infty}^{\infty} g^*(\hat{x}) \exp(-2 \pi i \omega \hat{x}) d\hat{x}
    \end{aligned}
\end{equation}
where the latter equation represents the Fourier transform of the wavelet $g$ over the short domain $\hat{x} \in x$.
Within the WNO framework, a wavelet-based integral operator governs function evolution across wavelet integral layers. Specifically, for a higher dimensional representation $v_j(x) = \wp(\xi)$ at the $j$-th wavelet integral layer, the transformation is given as,
\begin{equation}
v_{j+1}(x) = \sigma\left( \left(\mathcal{K}_\phi * v_j\right) + b v_j \right)(x),
\end{equation}
where $\sigma \in \mathbb{R} \mapsto \mathbb{R}$ is a pointwise nonlinearity, $b \in \mathbb{R}$ is a learnable functional bias, and $\mathcal{K}_\phi$ is a kernel operator parameterized by $\phi \in \bm{\theta}$. 
The convolution $(\mathcal{K}_\phi * v_j)$ is implemented in the wavelet domain using the convolution theorem in wavelet space rather than convolution in the physical domain. 
The previous formulation of wavelet domain convolution implicitly assumes a simplified, elementwise product structure in the wavelet domain. In particular, the convolution is performed as $(\mathcal{K}_\phi * v_j) = \mathcal{W}^{-1}(\mathcal{W}(v_j) \cdot R_\phi)$, where $R_\phi$ is the kernels of the neural operator. 
This is only an approximation of the true wavelet domain convolution operator, since in the case of wavelet transforms, convolution does not reduce to pointwise multiplication, as it does under Fourier transforms, but instead retains its integral form.
This forces the network to learn the integral transform as well as the features of the underlying solution operator, creating an additional burden for the network.
Consequently, the formulation fails to capture the true underlying structure of convolution, particularly the interactions across different spatial locations and scales. 

As one of the novelties of this work, we exploit the true wavelet domain convolution structure and replace the simple elementwise product of the previous version with a more sophisticated convolution integral in the wavelet domain. 
In particular, we first represent the spatial convolution $h(x) = \mathcal{K}_\phi(x) \ast v_j(x)$ in the wavelet domain as $h(x) = \mathcal{W}^{-1}((\mathcal{W}h)(x))$, where $(\mathcal{W}h)(s,t) = \mathcal{W}(\mathcal{K}_\phi(x) \ast v_j(x))$ given the scale and translation parameters $s \in \mathbb{R}$ and $t \in \mathbb{R}$ is defined as,
\begin{equation}
\begin{aligned}
(\mathcal{W}h)(s,t)
    \;&=\;
    \int_{\Omega} \!
    \Bigl(\mathcal{K}_\phi(x) \ast v_j(x)\Bigr)
    \frac{1}{\sqrt{|s|}}\,
    g\!\Bigl(\tfrac{x - t}{s}\Bigr)\,dx, \\
    &=\;
    \int_{\Omega} \!
    \Bigl(\!\int_{\Omega} v_j(x-\tau)\, \mathcal{K}_\phi(\tau)\,d\tau\Bigr)
    \frac{1}{\sqrt{|s|}}\,
    g\!\Bigl(\tfrac{x - t}{s}\Bigr)\,dx.
\end{aligned}
\end{equation}
Assuming the functions involved are sufficiently regular (e.g., square-integrable) and the energy under the integrals is finite, we apply Fubini's Theorem \cite{debnath2015wavelet} to interchange the order of integration, we obtain,
\begin{equation}
    \begin{aligned}
        (\mathcal{W}h)(s,t)
        \;&=\;
        \int_{\Omega} \mathcal{K}_\phi(\tau)\,
        \underbrace{\biggl[
            \frac{1}{\sqrt{|s|}}
            \int_{\Omega}  v_j(x-\tau)\,
            g\!\Bigl(\tfrac{x - t}{s}\Bigr)\,dx
        \biggr]}_{\displaystyle (\mathcal{W}v_j)(s, t - \tau)} d\tau, \\
        &=\; \int_{\Omega} \mathcal{K}_\phi(\tau)\,
        (\mathcal{W}v_j)(s, t - \tau) d\tau.
    \end{aligned}
\end{equation}
where $(\mathcal{W}v_j)(s, t - \tau)$ denotes the wavelet transform of the input $v_j(x)$. 
In this paper, we propose an improved WNO architecture, referred to as enhanced WNO, that relaxes the pointwise multiplication assumption inherent within the original WNO with the above convolution structure in the wavelet domain.

\begin{figure}[ht!]
    \centering
    \includegraphics[width=\linewidth]{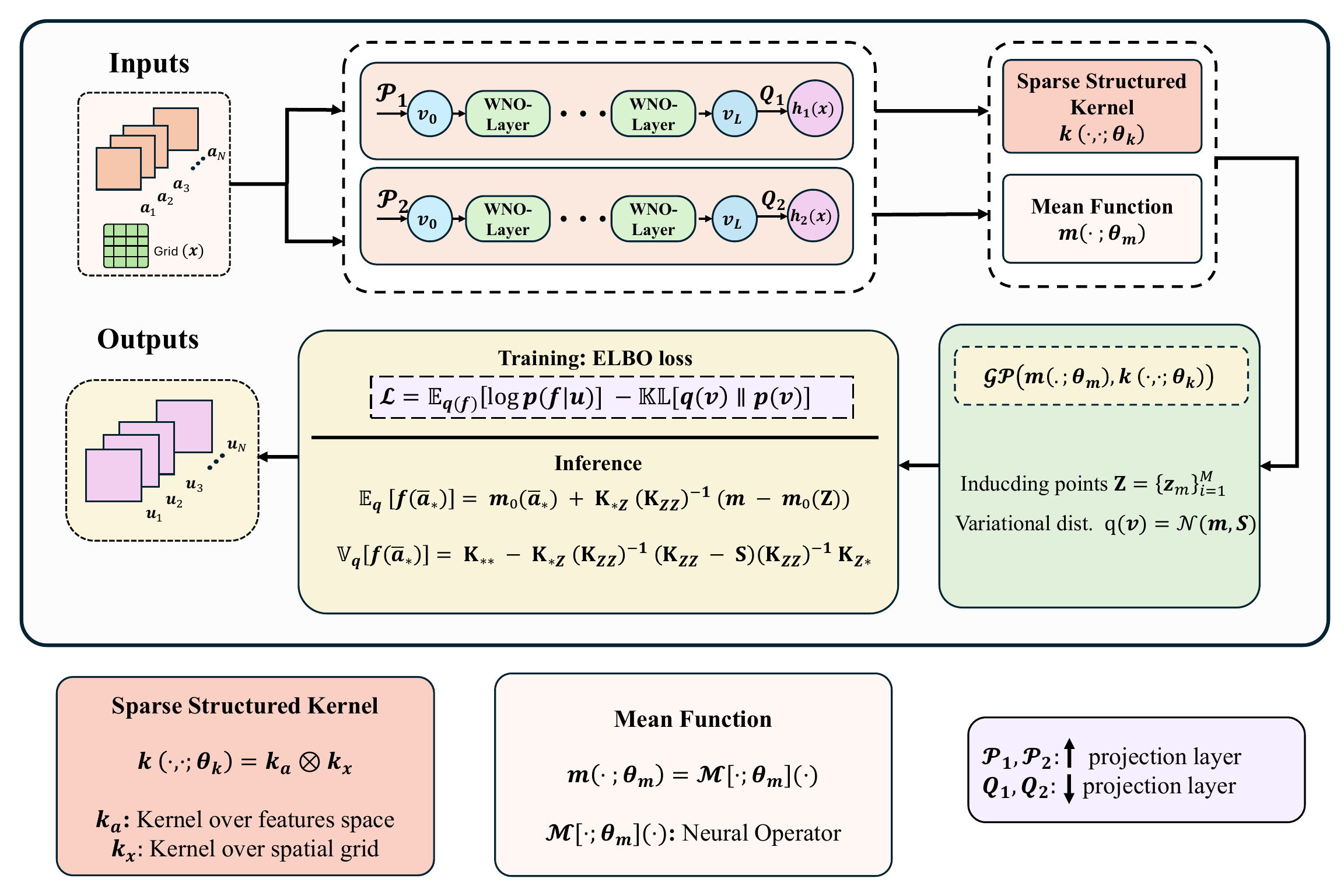}
    \caption{\textbf{LoGoS-GPO architecture.} Schematic representation of the proposed architecture for the LoGoS-GPO. The input field \( a(x) \) is processed through two distinct uplift projection layers, denoted by \( \mathcal{P}_1 \) and \( \mathcal{P}_2 \), which map the original input into high-dimensional latent representations suitable for the GP \emph{mean} and \emph{kernel} modeling paths, respectively. These projected embeddings are then passed through enhanced Wavelet Neural Operators (WNOs), producing latent representations \( h_1 \) and \( h_2 \). Here, \( h_1 \) captures rich operator-aware structures to serve as the non-stationary prior mean function \( m_0(\cdot) \), while \( h_2 \) encodes a latent representation from which the covariance kernel \( \mathbf{K}(\cdot,\cdot) \) is constructed. 
    Following this, these projected features, which are the latent outputs \( h_1 \) and \( h_2 \), inform a LoGoS-GPO with inducing points \( \mathbf{Z} \), enabling scalable variational inference using a learned variational distribution \( q(\mathbf{v}) \). The predictive posterior distribution yields the posterior mean \( \mathbb{E}_q[\bm{f}(\bar{\bm a}_*)] \) and variance \( \mathbb{V}_q[\bm{f}(\bar{\bm a}_*)] \), preserving uncertainty calibration. This dual-path encoding framework effectively decouples the learning of mean and kernel functions while maintaining end-to-end trainability.}
    \label{fig:schematics_sgp}
\end{figure}

While this convolution can be performed directly using local kernels like the convolutional neural networks (CNNs), they are insufficient for capturing the global dependencies required in operator learning tasks.
An alternative is to perform the convolution in the Fourier domain using elementwise multiplication. 
The motivation for using Fourier domain convolution is twofold: first, it provides a more principled and mathematically rigorous interpretation by avoiding the approximation inherent in the previous formulation; and second, it introduces a global convolution mechanism in the frequency domain, which naturally complements the localized, multiscale representation present in wavelets.  
Given that the network kernel $R_{\phi}$ is directly defined in the Fourier domain (thus reducing computational time associated with the Fourier transform of the kernel), the convolution in the Fourier domain is defined as,
\begin{equation}
    \int_{\Omega} \mathcal{K}_\phi(\tau)\,
        (\mathcal{W}v_j)(s, t - \tau) d\tau = \mathcal{F}^{-1}\!\Bigl[
    R_{\phi} 
    \;\cdot\; 
    \mathcal{F}\!\bigl(\mathcal{W}v_j\bigr)(s,\omega)
\Bigr],
\end{equation}
where $(\cdot)$ denotes the elementwise multiplication. Since the transformed output is in the wavelet space, the inverse wavelet transform is applied to the output to reconstruct the solution, yielding the following equation,
% up to appropriate scaling factors. Consequently,
\begin{equation}
\bigl(\mathcal{K}_{\phi} * v_j\bigr)(x)
\;=\;
\mathcal{W}^{-1}\!\Bigl(\,
    \mathcal{F}^{-1}\bigl[
        R_{\phi} 
        \,\cdot\, 
        \mathcal{F}(\mathcal{W}v_j)
    \bigr]
\Bigr)(x).
\end{equation}
The above equation shows that our modified wavelet convolution is essentially the 
\emph{inverse wavelet transform of output of the Fourier domain convolution}. 
This emphasizes the close connection between wavelet-based filtering and the usual \emph{global} convolution in the frequency domain while still preserving the desirable \emph{local} and \emph{multiscale} features afforded by wavelets.
Hence, the final WNO update is written compactly as,
\begin{equation}
v_{j+1}(x) \;=\; \sigma\!\Bigl(
    \mathcal{W}^{-1}\!\Bigl(\,
    \mathcal{F}^{-1}\bigl[
        R_{\phi}
        \,\cdot\, 
        \mathcal{F}(\mathcal{W}v_j)
    \bigr]
\Bigr)
    \;+\;
    b \, v_j
\Bigr)(x).
\end{equation}
It is implicitly understood that the wavelet transform (and its inverse) is computed via discrete wavelet transform (DWT), while the Fourier (and its inverse) transform is computed using the fast Fourier transform (FFT) algorithms, efficiently utilizing both wavelet- and Fourier-domain approaches.
The schematic representation of the overall architecture is shown in Fig.~\ref{fig:schematics_sgp}.

% Inference
\begin{algorithm}[ht]
    \caption{Inference}
    \label{alg:inference}
    \begin{algorithmic}[1]
    
    \Require 
    Trained parameters $\Theta^{(T)} = \{\bm{\theta}_m,\;\bm{\theta}_\phi,\;\bm{Z},\;\bm{m},\;\mathbf{S}\}$; test input $\bar{\bm{a}}_* = (\bm{a}_*, \bm{x}_*)$
    
    \Ensure 
    Mean and covariance of predictive posterior $q\left(\bm f_*\right)$
    \State Compute the mean terms:  $\bm m_0(\bar{\bm{a}}_*),\; \bm m_0^{(Z)}$ \Comment{Eqs.~\eqref{eq:mean_A}, \eqref{eq:mean_Z}}
    \State Compute feature and spatial kernel Matrices:\\
    \qquad $\mathbf{K}_a \gets \mathbf{K}_{a}(\Phi_{\bm\theta_{k_3}}({\bm a}_*), \Phi_{\bm\theta_{k_3}}(\mathbf{Z})) $,\\
    \qquad $\mathbf{K}_x \gets \left[ k_x(\bm{x}_*, \bm{x}_j)\right]_{j \in \mathcal{S}}$
    \Comment{Eqs.~\eqref{eq:kf_kernel},\eqref{eq:NN_kernel}}
    \State Compute predictive mean and covariance: $\mathbb{E}[\bm f_*]$, $\mathrm{Var}[\bm f_*]$ 
    \Comment{Eq.~\eqref{eq:pred_moments}}
    
    \State \textbf{Return} Mean and covariance of $q\left(\bm f_*\right)$
    
    \end{algorithmic}
\end{algorithm}

\section{Numerical Examples}\label{Numercial_eg}
In this section, we illustrate a variety of case studies to test the performance of the proposed LoGoS-GPO. The numerical examples selected include both 1D and 2D problems with varied levels of difficulty (see Table \ref{table_summary_dataset}). We investigate the performance of the proposed approach in terms of accuracy and scalability. While the relative $L_2$ loss is used to assess the accuracy of the proposed approach, the wall-clock time and memory required are used for investigating the scalability of the proposed approach. We carried out a thorough investigation involving batch size, learning rate, level of wavelet decomposition, quasi-Newton optimizers, and number of epochs. The optimal settings obtained for each of the examples are shown in Table \ref{table_experimental_settings}.

\begin{table}[ht!]
\centering
\caption{Summary of datasets used in the case studies}
\label{table_summary_dataset}
\begin{threeparttable}
% \footnotesize 
\begin{tabular}{@{}llll@{}}
\toprule
\textbf{Example} & \textbf{Input} & \textbf{Output} & \textbf{Input Distribution} \\
\midrule
Burgers' & Initial condition & Solution at final time $T$ & Gaussian random field \\
Wave Advection    & Initial condition & Solution at $T = 0.5$        & Random square-parabolic wave \\
Darcy (triangular)  & Boundary condition  & Pressure field & Gaussian random field \\
Navier–Stokes     & Forcing term       & Vorticity at $T = 10$       & Gaussian random field \\
\bottomrule
\end{tabular}
\end{threeparttable}
\end{table}

\begin{table}[ht!]
\centering
\caption{Experimental settings for case studies}
\label{table_experimental_settings}
\begin{threeparttable}
\footnotesize
    \begin{tabular}{lcccccccc}
    \hline
    \textbf{Example} & \textbf{\# Training} & \textbf{\# Testing} & \textbf{Batch} & \textbf{*LWD} & \textbf{Learning} & \textbf{Optimizer} & \textbf{Epochs} \\
    \textbf{} & \textbf{samples} & \textbf{samples} & \textbf{size} & \textbf{} & \textbf{rate} & \textbf{} & \textbf{} \\
    \hline
    Burgers$^{(\ref{ne:burger})}$ & 2000 & 100 & 32 & 5 & $8\times10^{-3}$ & AdamW & 500 \\
    Wave advection$^{(\ref{ne:wa})}$ & 2000 & 100 & 32 & 3 & $2\times 10^{-2}$ & AdamW & 500 \\
    Darcy (notch)$^{(\ref{ne:darcy})}$ & 1500 & 100 & 16 & 3 & $4\times10^{-3}$ & AdamW & 500 \\
    Navier-Stokes$^{(\ref{ne:ns})}$ & 3000 & 100 & 16 & 4 & $4\times10^{-4}$ & LBFGS & 400 \\
    \hline
    \end{tabular}
    \begin{tablenotes}
        \item[*] LWD: Level of wavelet decomposition
    \end{tablenotes}
\end{threeparttable}
% \caption*{LWD*: Level of wavelet decomposition}
\end{table}

\begin{table}[ht]
\centering
\caption{Relative $L_2$ error between the ground truth and the predicted results for the test set.}
\label{table_accuracy}
\begin{threeparttable}
    \begin{tabular}{lccc}
    \hline
    \multirow{2}{*}{\textbf{Case studies}} & \multicolumn{3}{c}{\textbf{Frameworks}} \\ \cline{2-4}
     & \textbf{LoGoS-GPO} & \textbf{GPO} & \textbf{SVGP} \\
    \hline
    Burger & $\mathbf{0.86 \pm 0.07}\,\%$ & $2.89 \pm 0.42\,\%$ &  $3.81 \pm 0.09\,\%$\\
    Wave Advection & $\mathbf{0.43 \pm 0.11}\,\%$ & $0.63 \pm 0.21\,\%$ & $1.81 \pm 0.29\,\%$  \\
    Darcy (notch) & $\mathbf{1.38 \pm 0.21}\,\%$ & $2.18 \pm 0.29\,\%$ & $5.18 \pm 0.47\,\%$ \\
    Navier Stokes & $\mathbf{2.01 \pm 0.53}\,\%$ & $2.21 \pm 0.32\,\%$ &$7.89 \pm 0.22\,\%$  \\
    \hline
    \end{tabular}
\end{threeparttable}
\end{table}
%%%%

\subsection{Case study 1: 1D Burger equation}\label{ne:burger}
The 1D Burgers' equation is a fundamental partial differential equation frequently used to model nonlinear transport phenomena in areas such as fluid dynamics, gas dynamics, and traffic flow. A distinctive aspect of this equation is its ability to capture both advective (nonlinear convection) and diffusive behavior.
In this first numerical example, we consider the viscous form of the 1D Burgers' equation defined on a periodic domain, given by:
\begin{equation}\label{eq_BS}
\begin{aligned}
\partial_{t} u(x, t) + 0.5\,\partial_{x} \left(u^{2}(x,t)\right) &= \nu \,\partial_{x x} u(x,t), && x \in (0,1), \; t \in (0,1], \\
u(0, t) &= u(1, t), && t \in (0,1], \\
u(x, 0) &= u_{0}(x), && x \in (0,1),
\end{aligned}
\end{equation}
where \( \nu > 0 \) denotes the viscosity coefficient, and \( u_0(x) \) specifies the initial condition.
The initial velocity field \( u_0(x) \) is sampled from a Gaussian random field with zero mean and a covariance kernel defined by
\begin{equation}
u_o(x) \sim \mathcal{N}\left(0, \, 625(-\Delta + 25I)^{-2}\right),
\end{equation}
where \( \Delta \) represents the Laplace operator with periodic boundary conditions. This construction ensures spatially smooth initial conditions with a controllable correlation length scale.
The primary objective is to learn the solution operator that maps the initial state \( u(x, t=0) \) to the final state \( u(x, t=1) \). For this example, we set the viscosity \( \nu = 0.1 \) and discretize the spatial domain using \(1024\) grid points. The dataset used in our experiments is publicly available and follows the setup introduced in \cite{li2020fourier}.

\begin{figure}[ht!] 
	\centering 
        \includegraphics[width=\textwidth]{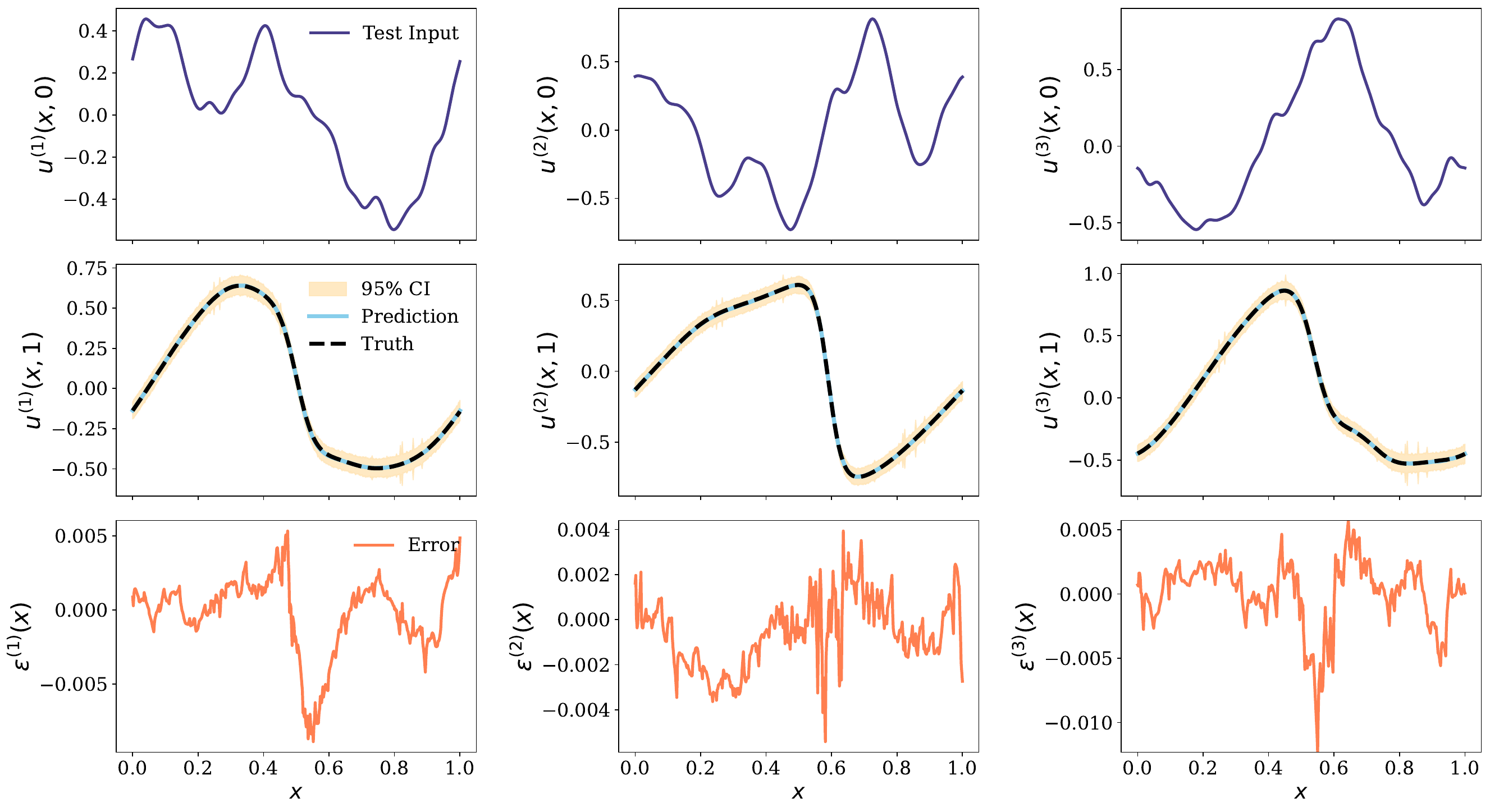} 
	\caption{\textbf{1D Burger}.  Figure illustrates the predictive performance of the proposed framework on three representative test inputs, each shown in a column. The first row displays the test input \( u^{\text{in}}(x, 0) \) over a spatial grid of resolution 1024. The second row compares the predicted solution \( \hat{u}^{\text{pred}}(x, 1) \) against the ground truth, with shaded regions indicating the 95\% confidence interval. The third row presents the pointwise prediction error \( \varepsilon(x) = \hat{u}^{\text{pred}}(x, 1) - u^{\text{true}}(x, 1) \).} 
	\label{fig:burger_pred} 
\end{figure}

\textbf{Results:} We train both LoGoS-GPO and GPO on a dataset of resolution \(1024\), evaluating their predictive performance on unseen discretized test function values. Fig.~\ref{fig:burger_pred} illustrates the predictive performance of our method. 
The predicted mean posterior exhibits minimal deviation from the ground truth, and the 95\% confidence intervals remain tight throughout the domain. The pointwise error plots further confirm the robustness of the proposed LoGoS-GPO, with errors consistently localized and within narrow bounds. 
Additionally, the performance of the proposed approach as opposed to vanilla GPO and Sparse Variation Gaussian Process (SVGP) is shown in Table \ref{table_accuracy}. The proposed LoGoS-GPO ($\epsilon \approx 0.86\%$) is found to outperform GPO ($\epsilon \approx 2.89\%$) and SVGP ($\epsilon \approx 3.81\%$) by a significant margin. 
These results validate the ability of our proposed framework to learn complicated input–output mappings while quantifying predictive uncertainty effectively.

\begin{figure}[ht!]
    \centering
    \begin{subfigure}[b]{\textwidth}
        \centering
        \includegraphics[width=\textwidth]{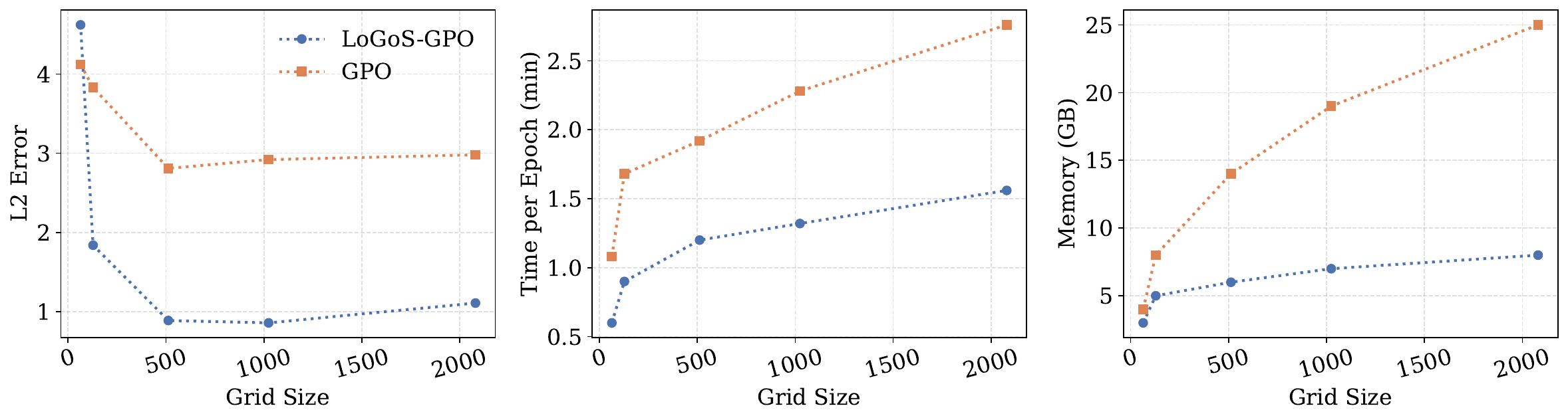}
        \caption{\textbf{Effect of grid resolution on model performance.} Comparison between LoGoS-GPO and GPO on the 1D Burger equation with a fixed training set size of 2000. \textbf{Left:} \(L2\) error across varying grid sizes. \textbf{Middle:} Wall clock time per epoch increases with resolution. However, computation is consistently faster with LoGoS-GPO. \textbf{Right:} Total memory usage during training and inference, showing a lower footprint for LoGoS-GPO.}
        \label{fig:burger_grid}
    \end{subfigure}
    
    \vspace{1em}

    \begin{subfigure}[b]{\textwidth}
        \centering
        \includegraphics[width=\textwidth]{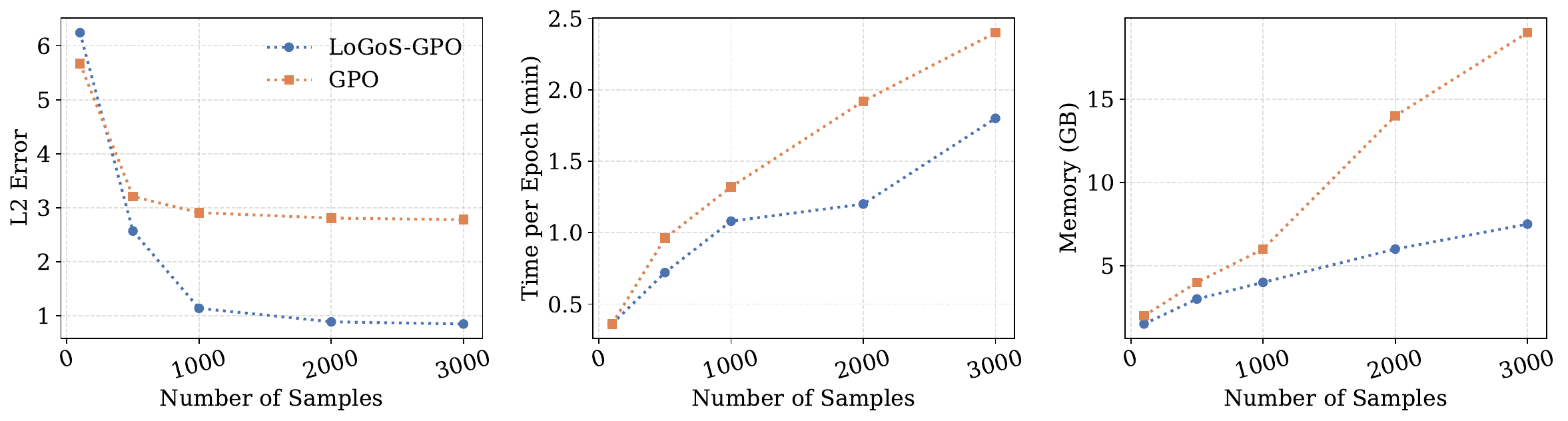}
        \caption{\textbf{Effect of training sample size on model performance.} Comparison for a fixed grid size of 512. \textbf{Left:} \(L2\) error vs. number of training samples. \textbf{Middle:} Wall-clock time per epoch as training sample size increases. \textbf{Right:} Memory consumption vs number of training samples, LoGoS-GPO maintains better scalability.
      }
        \label{fig:burger_samples}
    \end{subfigure}
    
    \caption{ \textbf{Scalability and performance comparison for 1D Burger Equation.} Figs.~\ref{fig:burger_grid} and~\ref{fig:burger_samples} compare LoGoS-GPO and GPO across varying grid resolutions and training sample sizes. LoGoS-GPO demonstrates improved efficiency in accuracy, wall clock time, and memory usage, demonstrating its suitability for large-scale problem settings.}
    \label{fig:burger_combined}
\end{figure}

We further assess the scalability and efficiency of LoGoS-GPO with an increase in resolution and number of training samples. The results obtained are shown 
in Fig.~\ref{fig:burger_combined}.
We observe that the proposed LoGoS-GPO has significantly better scalability with grid size as opposed to the vanilla GPO (Fig. \ref{fig:burger_grid}). Similar observation holds for an increase in the number of samples as well (Fig. \ref{fig:burger_samples}).
Additionally, we note that the accuracy deteriorates slightly with an increase in the grid size (Fig. \ref{fig:burger_grid}, leftmost plot). This is expected as the difficulty associated with the learning increases with an increase in resolution. Overall, the fact that the proposed LoGoS-GPO exhibits better computational efficiency, with lower runtime and memory cost, underscores its advantage in large-scale settings

\subsection{Case study 2: 1D Wave advection equation}\label{ne:wa}
The wave advection equation is a hyperbolic PDE widely used in physics and engineering to model the transport of scalar quantities under a prescribed velocity field. It serves as a foundational tool for simulating wave propagation phenomena, such as sound waves in the atmosphere or surface waves in fluids. 
This second numerical example considers the 1D wave advection equation with periodic boundary conditions, capturing long-term behavior without boundary reflections. The governing PDE is given by:
\begin{equation}
    \begin{aligned}
    \partial_t u(x,t) + \nu \,\partial_x u(x,t) &= 0, && x \in (0,1),\; t \in (0,1), \\
    u(x - \pi) &= u(x + \pi), && x \in (0,1),
    \end{aligned}
\end{equation}
where \( \nu > 0 \) is the constant advection speed, and \( u(x,t) \) denotes the scalar quantity being transported. The domain is defined over the interval \( (0,1) \) with periodic boundary conditions to mimic a continuous medium.
The initial condition is defined as a superposition of a square wave and a smooth parabolic bump:
\begin{equation}\label{wave_init}
    u(x,0) = h\,\mathbb{I}_{\left\{c - \frac{\omega}{2},\, c + \frac{\omega}{2} \right\}}(x) + \sqrt{ \max\left(h^2 - (a(x - c))^2, \, 0\right) }.
\end{equation}
Here, \( \omega \) controls the width and \( h \) the height of the square wave. The indicator function \( \mathbb{I}_{\left\{c - \frac{\omega}{2},\, c + \frac{\omega}{2} \right\}}(x) \) equals 1 when \( x \in \left[c - \frac{\omega}{2},\, c + \frac{\omega}{2} \right] \) and 0 otherwise. The wave is centered at position \( x = c \), and the parameters \( (c, \omega, h) \) are independently sampled from the ranges \( c \in [0.3, 0.7] \), \( \omega \in [0.3, 0.6] \), and \( h \in [1, 2] \), respectively.

\begin{figure}[ht!] 
	\centering 
        \includegraphics[width=\textwidth]{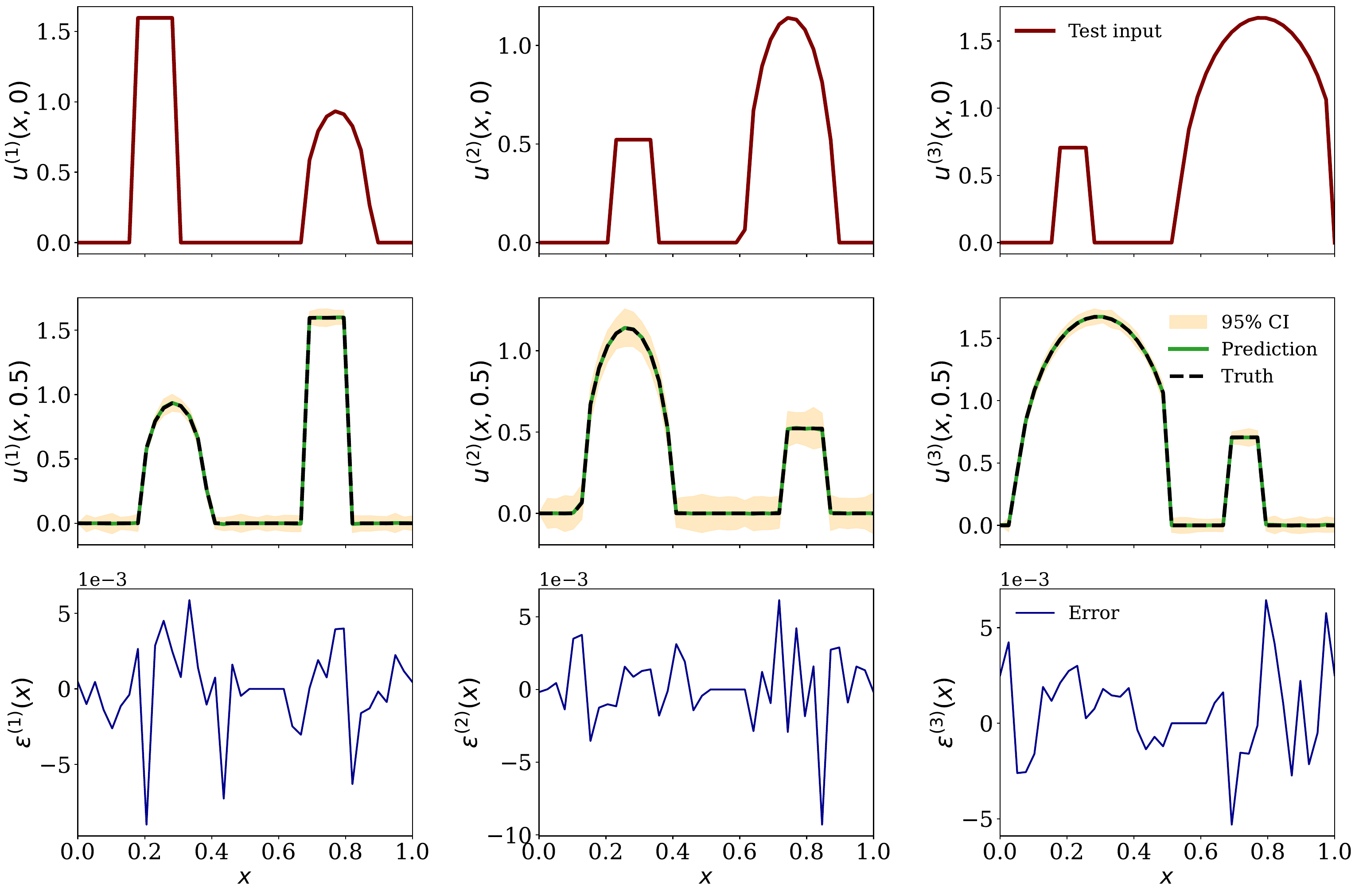} 
	\caption{\textbf{1D Wave advection equation}. Figure shows the prediction obtained from the proposed framework LoGoS-GPO. The first row shows one of the three representative test inputs on a spatial resolution of 200. The second row shows the corresponding ground truth and the mean prediction obtained from our proposed framework, along with a 95 \% confidence interval. The third row shows the pointwise error between the ground truth and the mean predictions.} 
	\label{fig:wa_pred} 
\end{figure}

\textbf{Results:} For this example, we evaluate the predictive capability of the proposed LoGoS-GPO framework on a dataset of spatial resolution \(200\). Fig.~\ref{fig:wa_pred} presents predictions on a set of test functions drawn from the input distribution. The LoGoS-GPO successfully captures the sharp transitions and flat segments characteristic of the wave advection equation, with the predicted mean solution closely tracking the ground truth across the domain. The confidence bands remain narrow, even near discontinuities, reflecting strong uncertainty calibration. Quantitative performance metrics are summarized in Table~\ref{table_accuracy}, and the corresponding hyperparameter settings are listed in Table~\ref{table_experimental_settings}. The low-magnitude pointwise errors in the third row further indicate the stability and precision of the LoGoS-GPO in approximating the underlying mapping.

\begin{figure}[ht!]
    \centering
    \begin{subfigure}[b]{\textwidth}
        \centering
        \includegraphics[width=\textwidth]{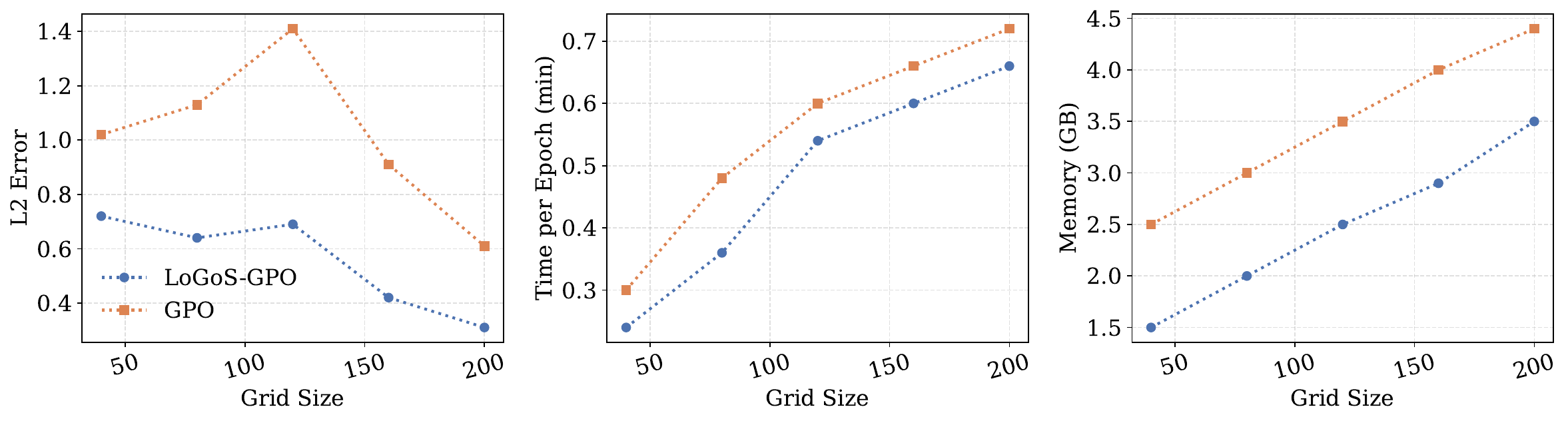}
        \caption{\textbf{Effect of grid resolution on model performance.} Comparison between LoGoS-GPO and GPO on the 1D Wave Advection problem with fixed training sample size of 2000. \textbf{Left:} \(L2\) error as a function of grid size, highlighting improved accuracy of LoGoS-GPO across resolutions. \textbf{Middle:} Wall-clock time per epoch increases with grid size, with LoGoS-GPO remaining consistently faster. \textbf{Right:} Memory footprint for both training and inference, showing reduced memory usage for LoGoS-GPO}
        \label{fig:wa_grid}
    \end{subfigure}
    
    \vspace{1em}

    \begin{subfigure}[b]{\textwidth}
        \centering
        \includegraphics[width=\textwidth]{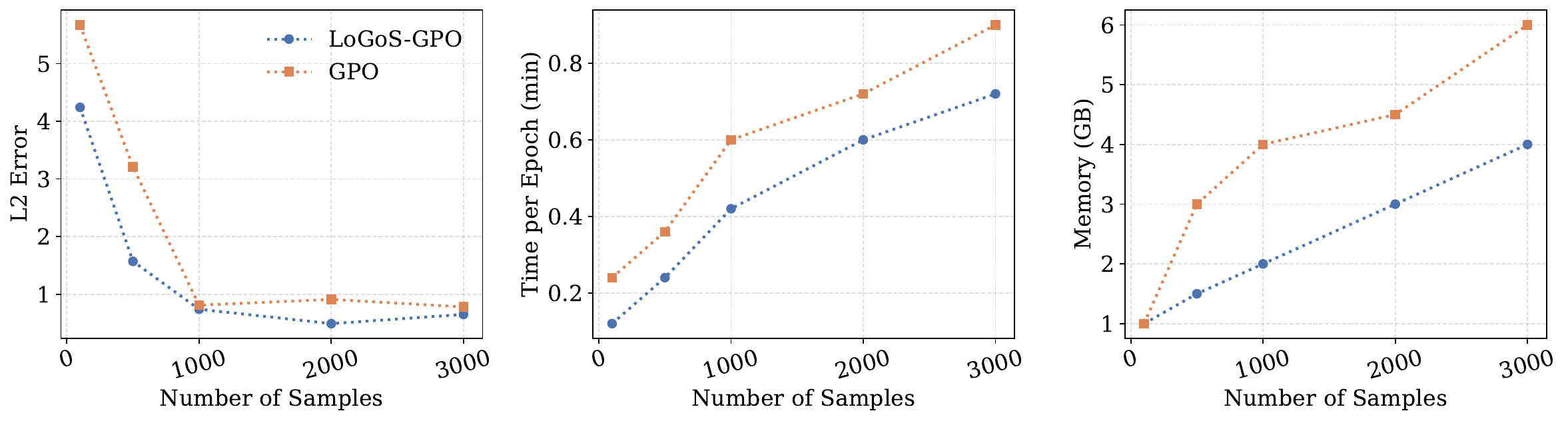}
        \caption{\textbf{Effect of training sample size on model performance.} Comparison between LoGoS-GPO and GPO for a fixed grid size of 200. \textbf{Left:} \(L2\) error decreases with sample size for both models, with LoGoS-GPO showing lower error at large sample sizes. \textbf{Middle:} Wall clock time per epoch scales almost-linearly for LoGoS-GPO compared to GPO. \textbf{Right:} Memory usage increases with sample size, and remains lower for LoGoS-GPO.}
        \label{fig:wa_samples}
    \end{subfigure}
    
    \caption{\textbf{Scalability and performance analysis for 1D wave advection equation.} We evaluate the LoGoS-GPO and GPO across varying grid sizes (Fig.~\ref{fig:wa_grid}) and training sample sizes (Fig.~\ref{fig:wa_samples}). LoGoS-GPO consistently demonstrates lower error, faster training time, and reduced memory footprint, validating its effectiveness for scalable operator learning problems.}
    \label{fig:wa_combined}
\end{figure}

To evaluate computational performance, we further conduct scalability experiments summarized in Fig.~\ref{fig:wa_combined}. In Fig.~\ref{fig:wa_grid}, we vary the grid resolution while fixing the number of training samples, and in Fig.~\ref{fig:wa_samples}, we vary the sample size for a fixed resolution. LoGoS-GPO demonstrates favorable scaling trends: it consistently outperforms the vanilla GPO in runtime and memory consumption while preserving accuracy. These results highlight the suitability of LoGoS-GPO predictive uncertainty, and computational efficiency is crucial.

\subsection{Case study 3: 2D Darcy's equation in the triangular domain}\label{ne:darcy}
\begin{figure}[ht!] 
	\centering 
        \includegraphics[width=\textwidth]{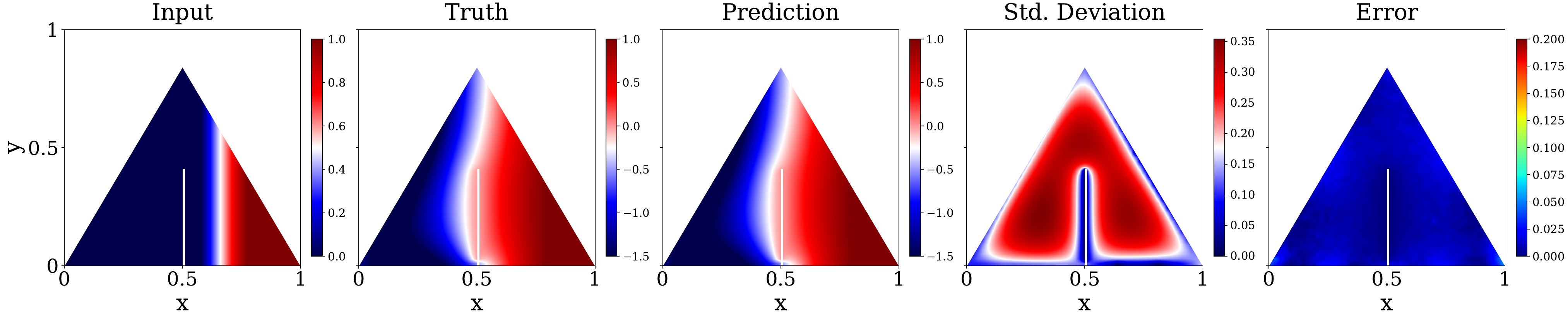} 
	\caption{\textbf{2D Darcy flow with a notch in the triangular domain.} Figure illustrates the predictive performance of the proposed framework on a representative test sample over a spatial domain discretized with a \(51 \times 51\) grid. Each column represents a different subplot: \textbf{Input} shows the one boundary condition; \textbf{Truth} depicts the ground truth pressure field; \textbf{Prediction} shows the mean of the predicted pressure field; \textbf{Uncertainty} represents the standard deviation from the predictive posterior, indicating regions of predictive uncertainty; and \textbf{Error} shows the absolute error between the predicted mean and the ground truth. The model captures the underlying structure with excellent accuracy.
} 
	\label{fig:darcy_n_pred} 
\end{figure}

For the third case study, we consider the Darcy flow equation, which is an elliptic partial differential equation commonly used to model fluid transport through porous media. It plays a central role in various fields, such as hydrology, petroleum engineering, and subsurface flow modeling. In this example, we consider a 2D Darcy flow problem defined on a triangular domain with a notch, presenting a geometrically challenging setup. The governing equation for the steady-state Darcy flow is given by:
\begin{equation}
\begin{aligned}
- \nabla \cdot (a(x,y) \nabla u(x,y)) &= f(x,y), && (x, y) \in \Omega, \\
u(x,y)\big|_{\partial \Omega} &= g(x,y), && (x, y) \in \partial \Omega,
\end{aligned}
\end{equation}
where \( u(x, y) \) denotes the pressure field and \( a(x, y) \) is the permeability of the porous medium. The domain \( \Omega \subset \mathbb{R}^2 \) is a unit triangle with an interior notch, making the geometry non-trivial. The forcing function is set to a constant value \( f(x, y) = -1 \), and the permeability is fixed at \( a(x, y) = 0.1 \).
In this setup, the Dirichlet boundary condition is not fixed but instead modeled as a sample from a Gaussian process prior:
\begin{equation}
\begin{aligned}
u(x, y)\big|_{\partial \Omega} &\sim \mathcal{GP}(0,\, k((x, y), (x', y'))), \\
k((x, y), (x', y')) &= \exp\left( -\left( \frac{(x - x')^2}{2 l_x^2} + \frac{(y - y')^2}{2 l_y^2} \right) \right),
\end{aligned}
\end{equation}
where \( l_x = l_y = 0.2 \) are the length-scale parameters of the squared exponential kernel.
The goal in this numerical experiment is to learn the mapping from the boundary condition values to the corresponding pressure solution in the interior of the domain, i.e.,
\[
u(x,y)\big|_{\partial \Omega} \mapsto u(x,y), \quad (x,y) \in \Omega.
\]
\textbf{Results:} In the third case study, we consider the 2D Darcy flow equation defined over a triangular domain with a sharp notch, posing a challenge due to the geometric discontinuity and nontrivial boundary behavior. The proposed LoGoS-GPO is trained on dataset discretized over a grid resolution of \(51 \times 51\), with predictive evaluation on a set of unseen test functions. Fig.~\ref{fig:darcy_n_pred} displays the pressure field prediction for one representative test case. The LoGoS-GPO predicts the pressure field, even around the notch where sharp transitions and gradients are observed. The uncertainty captured via the posterior standard deviation provides an estimate of the predictive uncertainty in the form of predictive uncertainty. The corresponding error plot confirms the accuracy of the model, with minimal deviation from the ground.

\begin{figure}[ht!]
    \centering
    \begin{subfigure}[b]{\textwidth}
        \centering
        \includegraphics[width=\textwidth]{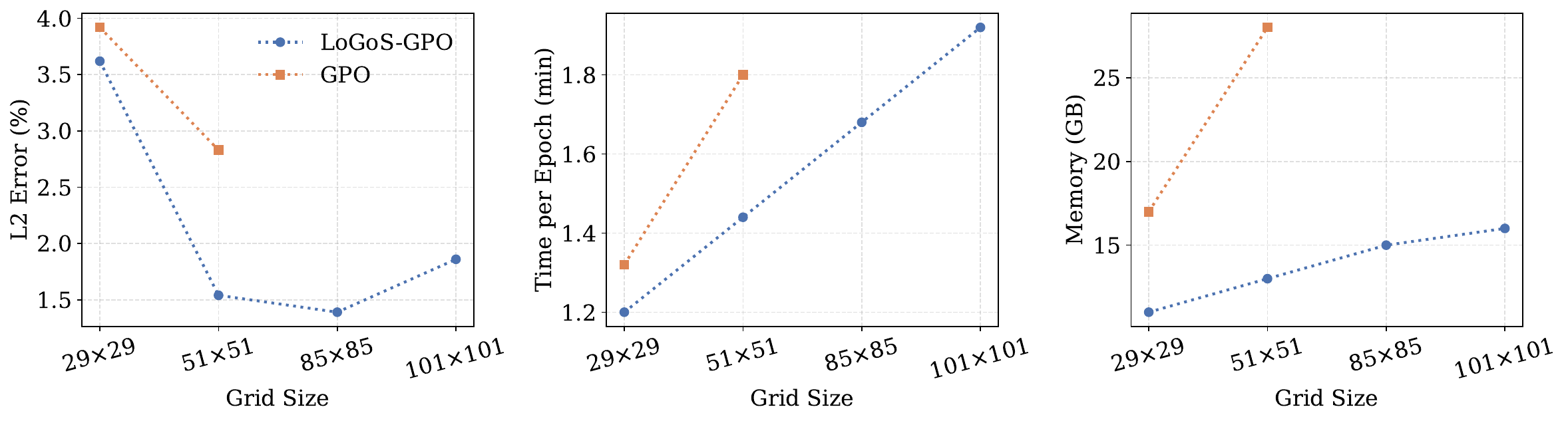}
        \caption{\textbf{Effect of grid resolution on model performance.} Comparison between LoGoS-GPO and GPO on the 2D Darcy flow equation with a fixed training sample size of 1500. \textbf{Left:} \(L2\) error across varying grid sizes. \textbf{Middle:} Wall clock time vs grid sizes. It clearly shows that the wall call time increases per epoch with increasing resolution. However, computation is consistently faster with LoGoS-GPO. \textbf{Right:} Total memory usage during training and inference, showing a lower footprint for LoGoS-GPO}
        \label{fig:darcy_n_grid}
    \end{subfigure}
    
    \vspace{1em}

    \begin{subfigure}[b]{\textwidth}
        \centering
        \includegraphics[width=\textwidth]{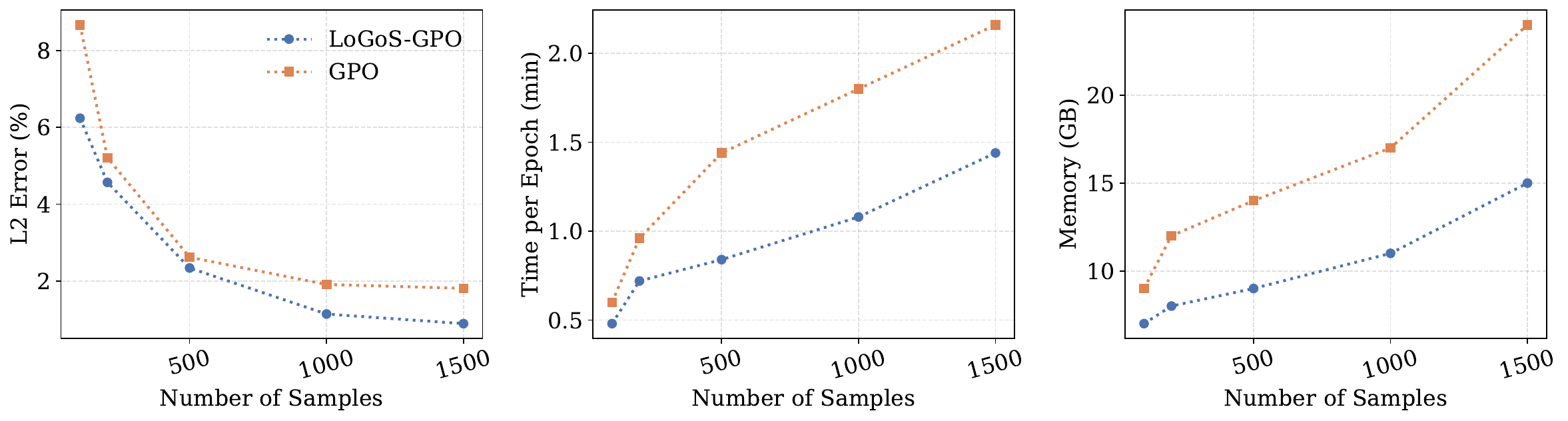}
        \caption{\textbf{Effect of training sample size on model performance.}Comparison for a fixed grid size of \(51\times 51\). \textbf{Left:} \(L2\) error vs. number of training samples. \textbf{Middle:} Wall-clock time per epoch as training sample size increases. \textbf{Right:} Memory consumption during training and inference, LoGoS-GPO maintains better scalability.
      }
        \label{fig:darcy_n_samples}
    \end{subfigure}
    
    \caption{ \textbf{Scalability and performance comparison for 2D Darcy flow equation with a notch in a triangular domain.} Figs.~\ref{fig:darcy_n_grid} and~\ref{fig:darcy_n_samples} compare LoGoS-GPO and GPO across varying grid resolutions and training sample sizes. LoGoS-GPO demonstrates improved efficiency in accuracy, wall clock time, and memory usage, demonstrating its suitability for fine grid problems.}
    \label{fig:darcy_n_combined}
\end{figure}

Scalability and performance results for this 2D problem are presented in Fig.~\ref{fig:darcy_n_combined}. When varying spatial resolution (Fig.~\ref{fig:darcy_n_grid}) and training sample size (Fig.~\ref{fig:darcy_n_samples}), the LoGoS-GPO consistently achieves lower computational overhead compared to the GPO. Notably, both memory usage and runtime remain significantly lower even as the resolution or dataset size increases. This efficiency arises because LoGoS-GPO exploits Kronecker decomposition over the spatial grid in 2D settings, substantially reducing computational cost. Meanwhile, the accuracy remains competitive or improves, especially in finer discretizations. These observations demonstrate the effectiveness of LoGoS-GPOs for operator learning problems, even for PDEs with irregular domains, offering a balance between accuracy and scalability.

\subsection{Case study 4: 2D Navier stokes equation}\label{ne:ns}
In the final case study, we consider the 2D incompressible Navier–Stokes equations, which play a fundamental role in modeling fluid motion in diverse physical systems, including atmospheric dynamics, ocean circulation, and aerodynamic flows. We adopt the vorticity-stream function formulation to simplify the formulation while preserving the essential dynamics. The governing equations are given by:
\begin{equation}
\begin{aligned}
\frac{\partial \omega}{\partial t} + (c \cdot \nabla)\omega - \nu \Delta \omega &= f', \quad \omega = -\Delta \xi, \\
\int_{\Omega} \xi &= 0, \quad c = \left( \frac{\partial \xi}{\partial x_2}, -\frac{\partial \xi}{\partial x_1} \right),
\end{aligned}
\end{equation}
where \( \omega \) denotes the vorticity, \( \xi \) is the stream function, \( \nu \) is the kinematic viscosity, and \( c \) is the velocity field derived from \( \xi \). The domain is periodic and defined over \( [0, 2\pi]^2 \).
The forcing term \( f' \) is modeled as a centered Gaussian random field with covariance operator
\[
\mathbf{C} = (-\Delta + 9\mathbf{I})^{-4},
\]
where \( \Delta \) is the Laplacian operator and \( \mathbf{I} \) is the identity matrix. The mean of the forcing term is zero. The initial vorticity field \( \omega(\cdot, 0) \) is fixed and sampled from the same Gaussian distribution. We set the viscosity to \( \nu = 0.025 \) and integrate the system until the final time \( T = 10 \). The objective in this example is to learn the mapping from the forcing term \( f' \) to the resulting vorticity field at the final time, that is \(f' \mapsto \omega(\cdot, T).\)

\begin{figure}
    \centering
    \includegraphics[width=\linewidth]{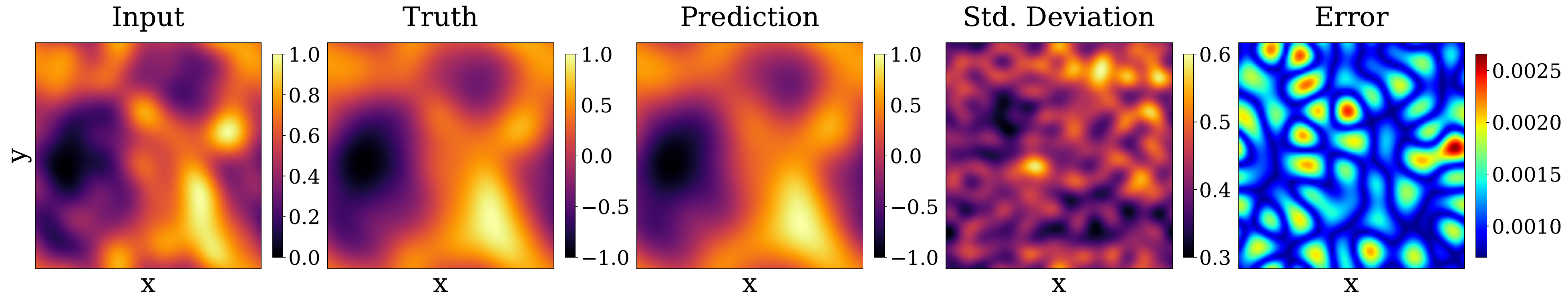}
    \caption{\textbf{2D Navier stokes equation.} Figure illustrates the predictive performance of the proposed framework on a representative test sample over a spatial domain discretized with a \(64 \times 64\) grid. Each column represents a different subplot: \textbf{Input} shows the one forcing term; \textbf{Truth} depicts the ground truth vorticity field; \textbf{Prediction} shows the mean of the predicted vorticity field; \textbf{Uncertainty} represents the standard deviation from the predictive posterior, indicating regions of predictive uncertainty; and \textbf{Error} shows the absolute error between the predicted mean and the ground truth.}
    \label{fig:ns_mean_pred}
\end{figure}

\textbf{Results:} As the last numerical example we consider 2D incompressible Navier–Stokes equations as a benchmark for evaluating the capability of our proposed framework. The proposed LoGoS-GPO is trained on vorticity fields discretized on a \(64 \times 64\) grid. Fig.~\ref{fig:ns_mean_pred} shows the mean predictions for a sample test case, highlighting the ability of LoGoS-GPO to recover the spatial structure of Navier Stokes flow. The predicted vorticity fields align closely with the ground truth, while the uncertainty map effectively captures regions of ambiguity. Importantly, the spatial distribution of predictive error remains low and is concentrated in areas with sharp vorticity gradients or complex interactions. The performance metrics and results are summarized in Table~\ref{table_accuracy}, with hyperparameters reported in Table~\ref{table_experimental_settings}.

\begin{figure}[ht!]
    \centering
    \begin{subfigure}[b]{\textwidth}
        \centering
        \includegraphics[width=\textwidth]{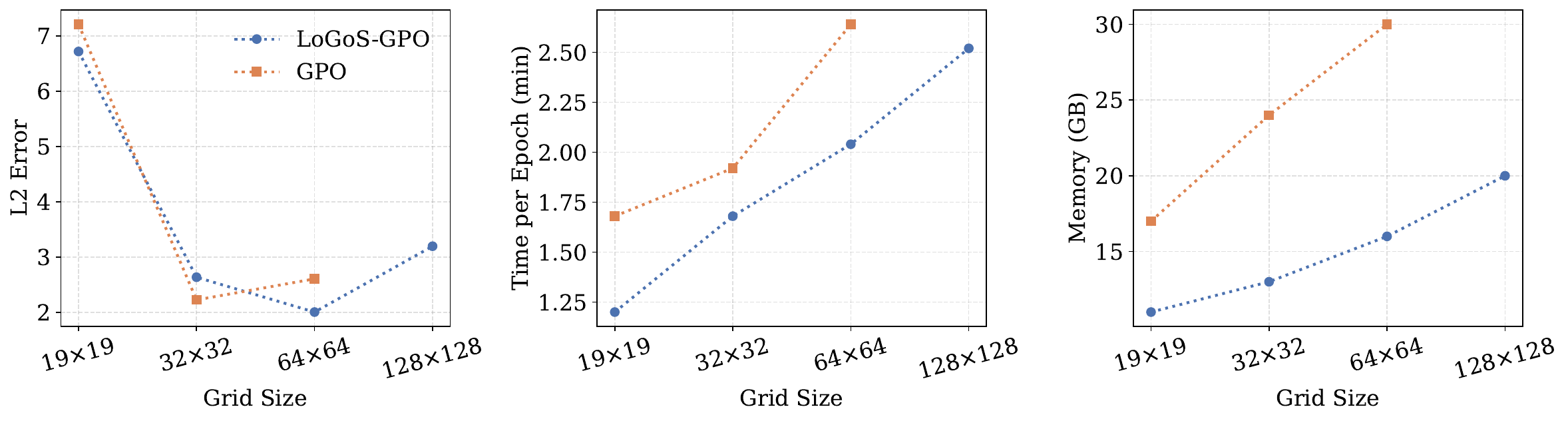}
        \caption{\textbf{Effect of grid resolution.}. Comparison between LoGoS-GPO and GPO on the 2D Navier-Stokes problem with a fixed training set size of 2000. \textbf{Left:} \(L2\) error across different grid sizes. \textbf{Middle:} Wall-clock time per epoch as a function of resolution. \textbf{Right:} Total memory usage during training and inference. LoGoS-GPO consistently reduces memory and runtime while maintaining competitive accuracy.} 
        \label{fig:ns_grid}
    \end{subfigure}
    
    \vspace{1em}

    \begin{subfigure}[b]{\textwidth}
        \centering
        \includegraphics[width=\textwidth]{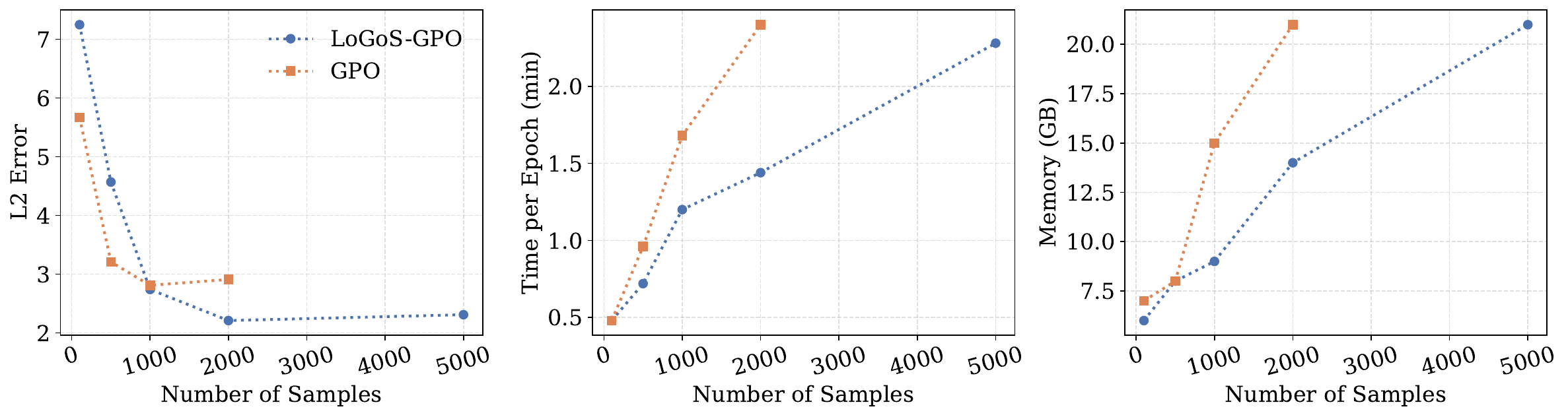}
        \caption{\textbf{Effect of training sample size on model performance.} Comparison at a fixed grid size of \(64 \times 64\). \textbf{Left:} \(L2\) error as the number of training samples increases. \textbf{Middle:} Wall-clock time per epoch. \textbf{Right:} Memory footprint during training and inference. LoGoS-GPO demonstrates better scalability with sample size, sustaining lower computational and memory overhead compared to GPO.} 
        \label{fig:ns_samples}
    \end{subfigure}
    
    \caption{\textbf{Scalability and performance comparison for 2D Navier Stokes equation: (a) Grid configuration; (b) Sampling points for training.} Figs. (a) and (b) compare LoGoS-GPO and vanilla GPO across varying spatial resolutions and sample sizes for the 2D Navier-Stokes problem. LoGoS-GPO achieves a favorable balance between accuracy and efficiency, making it suitable for solving complicated nonlinear PDEs in an uncertainty-aware manner.}
    \label{fig:ns_combined}
\end{figure}

The scalability of our method is analyzed in Fig.~\ref{fig:ns_combined}. Fig.~\ref{fig:ns_grid} evaluates performance across different grid sizes at a fixed training size of 2000 samples, while Fig.~\ref{fig:ns_mean_pred} investigates sample size variation at a fixed \(64 \times 64\) resolution. In both settings, LoGoS-GPO maintains competitive or better accuracy compared to GPO, while significantly reducing wall-clock time and memory requirements. These results demonstrate the efficiency and flexibility of our proposed approach, particularly in large-scale problems where uncertainty quantification and computational scalability are crucial.

\section{Conclusion}\label{Conclusion}
In this work, we introduced LoGoS-GPO, a scalable and uncertainty-aware operator learning framework that unites the interpretability of Gaussian process models with the expressiveness of neural operators. The framework addresses key computational challenges associated with traditional Gaussian Process Operators by combining local kernel approximations in the spatial domain using the nearest neighbor strategy, sparsity in the parameter space through sparse variational approximation, and structured decompositions using Kronecker products. This design enables efficient training and inference while maintaining high accuracy and the ability to quantify uncertainty. The inclusion of neural operator informed mean functions further enhances the model's flexibility in capturing complex input and output relationships arising in high-dimensional partial differential equations.

Through comprehensive experiments on a diverse set of nonlinear partial differential equations, including the Burgers equation, wave advection, Darcy flow, and Navier-Stokes equations, LoGoS-GPO consistently achieved strong predictive performance across varying spatial discretizations and parameter settings. Unlike deterministic operator learning methods, our framework offers calibrated and spatially resolved uncertainty estimates, making it well suited for scientific and engineering applications that demand both accuracy and reliability. The method demonstrates robustness across different resolutions and scales, showing clear potential for real-world deployment in data-scarce or high-resolution environments.

Beyond performance, LoGoS-GPO contributes a conceptual advancement by bridging classical kernel-based learning with modern operator learning. By integrating sparsity, locality, structure, and neural representations into a single probabilistic model, it establishes a pathway toward scalable and interpretable operator learning. Future research may explore the integration of physics-informed priors to embed physical laws directly into the kernel design, or the use of deep Gaussian processes to model hierarchical and multiscale dependencies. These directions will further expand the capability of LoGoS-GPO to serve as a foundational tool for uncertainty-aware modeling in complex physical and computational systems.

\section*{Acknowledgements}
SK acknowledges the support received from the Ministry of Education (MoE) in the form of a Research Fellowship. SC acknowledges the financial support received from Anusandhan National Research Foundation (ANRF) via grant no. CRG/2023/007667. SC and RN acknowledge the financial support received from the Ministry of Port and Shipping via letter no. ST-14011/74/MT (356529). 

\section*{Code availability}
On acceptance, all the source codes to reproduce the results in this study will be made available to the public on GitHub by the corresponding author.

% \section*{References}
% \bibliographystyle{unsrt}  
% \bibliography{bibliography}

\end{document}